\documentclass{article} 

\usepackage{hyperref}
\usepackage{url}
\usepackage{graphicx}
\usepackage{natbib}
\usepackage{booktabs}
\usepackage{wrapfig}
\usepackage{amsmath,amsfonts,bm,bbm}
\usepackage{multirow}
\usepackage{pifont}
\usepackage{caption}
\usepackage{subcaption}
\usepackage{colortbl}
\usepackage{microtype}

\usepackage{amssymb}
\usepackage{mathtools}
\usepackage{amsthm}

\usepackage[accepted]{icml2024}

\usepackage[capitalize,noabbrev]{cleveref}

\theoremstyle{plain}

\theoremstyle{definition}

\theoremstyle{remark}

\NewDocumentCommand{\rot}{O{45} O{1em} m}{\makebox[#2][l]{\rotatebox{#1}{#3}}}%

\definecolor{dark-yellow}{HTML}{eeae00}
\definecolor{dark-green}{HTML}{47e11e}

\usepackage[textsize=tiny]{todonotes}

\icmltitlerunning{Hyperbolic Active Learning for Semantic Segmentation under Domain Shift}

\begin{document}

\twocolumn[
\icmltitle{Hyperbolic Active Learning for Semantic Segmentation under Domain Shift}

\icmlsetsymbol{equal}{*}

\begin{icmlauthorlist}
\icmlauthor{Luca Franco}{equal,iai}
\icmlauthor{Paolo Mandica}{equal,sap}
\icmlauthor{Konstantinos Kallidromitis}{pan}
\icmlauthor{Devin Guillory}{ucb}
\icmlauthor{Yu-Teng Li}{ucb}
\icmlauthor{Trevor Darrell}{ucb}
\icmlauthor{Fabio Galasso}{iai,sap}
\end{icmlauthorlist}

\icmlaffiliation{sap}{Sapienza University of Rome}
\icmlaffiliation{pan}{Panasonic North America}
\icmlaffiliation{ucb}{UC Berkeley}
\icmlaffiliation{iai}{ITALAI S.R.L.}

\icmlcorrespondingauthor{Luca Franco}{luca.franco@italailabs.com}
\icmlcorrespondingauthor{Paolo Mandica}{paolo.mandica@uniroma1.it}

\icmlkeywords{Active Learning, Hyperbolic Neural Networks, Domain Adaptation, Semantic Segmentation}

\vskip 0.3in
]



\printAffiliationsAndNotice{\icmlEqualContribution} 

\begin{abstract}

We introduce a hyperbolic neural network approach to pixel-level active learning for semantic segmentation. 
Analysis of the data statistics leads to a novel interpretation of the hyperbolic radius as an indicator of data scarcity.
In HALO (Hyperbolic Active Learning Optimization), for the first time, we propose the use of epistemic uncertainty as a data acquisition strategy, following the intuition of selecting data points that are the least known. The hyperbolic radius, complemented by the widely-adopted prediction entropy, effectively approximates epistemic uncertainty.
We perform extensive experimental analysis based on two established synthetic-to-real benchmarks, i.e.\ GTAV $\rightarrow$ Cityscapes and SYNTHIA $\rightarrow$ Cityscapes. Additionally, we test HALO on Cityscape $\rightarrow$ ACDC for domain adaptation under adverse weather conditions, and we benchmark both convolutional and attention-based backbones.
HALO sets a new state-of-the-art in active learning for semantic segmentation under domain shift and it is the first active learning approach that surpasses the performance of supervised domain adaptation while using only a small portion of labels (i.e., 1\%).\footnote{Code available at \url{https://github.com/paolomandica/HALO}.}

\end{abstract}

\section{Introduction}
\label{sec:intro}

Dense prediction tasks, such as semantic segmentation (SS), are important in applications such as self-driving cars, manufacturing, and medicine. However, these tasks necessitate pixel-wise annotations, which can incur substantial costs and time inefficiencies \citep{cityscapes}. Previous methods \citep{xie2022ripu,Vu_2019_CVPR,shin2021labor,shin2021pixelpick,ning2021multi}
have addressed this labeling challenge via domain adaptation, capitalizing on large source datasets for pre-training and domain-adapting with few target annotations \citep{diffdomains}. 
Most recently, active domain adaptation (ADA) has emerged as an effective strategy, i.e.\ annotating only a small set of target pixels in successive labelling rounds~\citep{ning2021multi}.
State-of-the-art (SOTA) ADA \citep{shin2021labor,wu2022d2ada,xie2022ripu} relies on the entropy of predicted pseudo-labels, which they define as prediction uncertainty, as the core strategy for active learning (AL) data acquisition.
In fact, prediction uncertainty correlates well with the likelihood of pixel classification mistakes, but it is only one of the factors of the overall model uncertainty, as we argue in this work.

Following extensive literature \citep{Depeweg2017, kendall2017, hullermeier21uncert, toro22}, we distinguish aleatoric and epistemic uncertainty, and we propose the second for the data acquisition strategy in active learning. Epistemic uncertainty is an indicator of the state of knowledge about the task. This uncertainty stems not only from inaccuracies, as identified by prediction uncertainty, but also from the information the model has been exposed to thus far, including the amount of data considered. In the domain adaptation task, the domain gap arises from the model's lack of understanding of the new domain data, akin to the definition of epistemic uncertainty. Building upon this intuition, we propose HALO (Hyperbolic Active Learning Optimization), a novel approach for active domain adaptation, where we introduce the use of epistemic uncertainty into the data acquisition strategy. Our in-depth analysis shows that the hyperbolic radius effectively estimates data scarcity, revealing it as a key component in the estimation of epistemic uncertainty. The combination of the radius with a complementary information signal such as prediction entropy offers a comprehensive estimate of epistemic uncertainty.

\begin{figure*}[t]
\centering
\includegraphics[width=\textwidth]  
{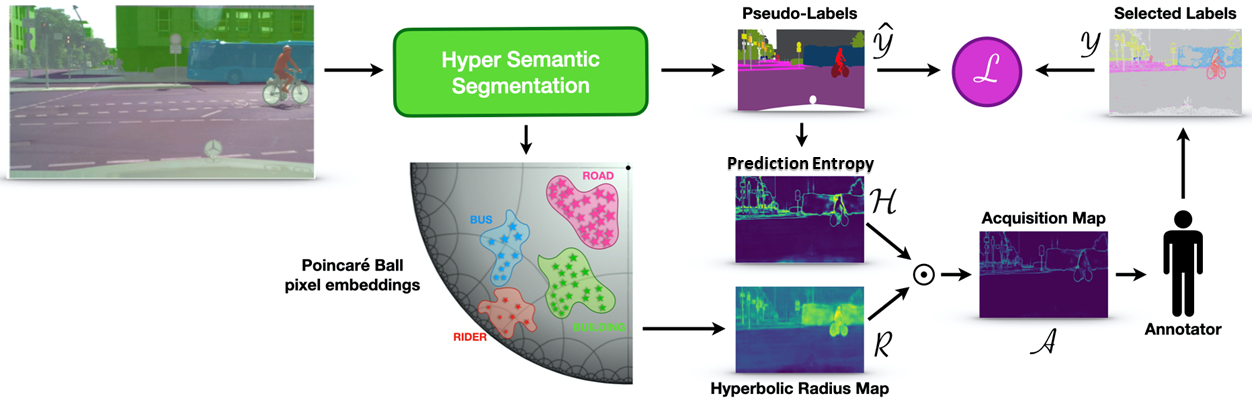}
\caption{
Overview of HALO. Pixels are encoded into the hyperbolic Poincaré ball and classified in the pseudo-label $\mathcal{\hat{Y}}$. The hyperbolic radius of the pixel embeddings defines the new hyperbolic score map $\mathcal{R}$. The prediction entropy $\mathcal{H}$ is extracted as the entropy of the softmax probabilities. Combining $\mathcal{R}$ and $\mathcal{H}$ we define the data acquisition score map $\mathcal{A}$, which is used to query new labels $\mathcal{Y}$. }
\label{fig:teaser}
\vspace{-0.2cm}
\end{figure*}

Interpreting the hyperbolic radius as a proxy to data scarcity diverges from known interpretations in the hyperbolic literature. The SOTA hyperbolic SS model ~\citep{atigh2022hyperbolic} trains with class hierarchies, which they manually define. As a result, their hyperbolic radius represents the parent-to-child hierarchical relations in the Poincaré ball. We adopt ~\citet{atigh2022hyperbolic} without enforcing hierarchical labels and we find that hierarchical relationships do not emerge naturally in our case. For instance, in HALO, classes such as \emph{road} and \emph{building} are closer to the center of the ball, while \emph{person} and \emph{rider} have larger radii.
This class arrangement contradicts the interpretation of the hyperbolic radius as a proxy for uncertainty, which emerged from metric learning hyperbolic studies~\citep{ermolov2022hyperbolic,franco23}. In our context, larger radii indicate larger data scarcity, therefore less certainty, which is in contrast with \citet{franco23}'s interpretation.
Thus, our interpretation of the hyperbolic radius as a proxy for data scarcity does not align with neither of the existing interpretations in the case of hierarchy-free hyperbolic SS.
Consider the HALO pipeline illustrated in Fig.~\ref{fig:teaser} and the circular sector representing the Poincaré ball, where pixels from various classes are mapped. The hyperbolic model assigns a higher radius to classes that appear less frequently in the dataset (e.g., \textit{rider}), and a lower radius to classes which are more frequent (e.g., \textit{road}). 
In Sec.~\ref{sec:interpretation}, we show how this novel interpretation of the hyperbolic radius arises bottom-up from data statistics.

We demonstrate the effectiveness of our approach through extensive benchmarking on well-established datasets for SS, including ADA from GTAV to Cityscapes, SYNTHIA to Cityscapes, and additional testing on Cityscapes to ACDC under adverse weather conditions.
HALO sets a new SOTA on all the benchmarks (+3.3\% on GTA$\rightarrow$CS, +4.2\% on SYNTHIA$\rightarrow$CS, and +2.9\% on CS$\rightarrow$ACDC). Moreover, this is the first AL method that surpasses the supervised domain adaptation baseline using only a small portion of labels (+2.6\% on GTA$\rightarrow$CS with 5\% budget).
Our paper also introduces a novel technique to enhance the stability of hyperbolic training, referred to as \textit{Hyperbolic Feature Reweighting} (HFR), cf.\ Sec.~\ref{sec:method}. Our code will be released.

In summary, our contributions include: 1) Presenting a novel interpretation of the hyperbolic radius as a proxy for data scarcity and its relationship with epistemic uncertainty; 2) Introducing hyperbolic neural networks in AL and a novel pixel-based data acquisition score based on the hyperbolic radius; 3) Validating both the concept and the algorithm through a comprehensive analysis, achieving a new state-of-the-art performance across all the considered ADA benchmarks for SS. Our method surpasses for the fist time in AL the supervised DA performance using only a small percentage of labels.

\section{Related Works}
\label{sec:rworks}

\textbf{Hyperbolic Representation Learning (HRL)} \quad
Hyperbolic geometry has been extensively used to capture embeddings of tree-like structures~\citep{nickel2017poincare,chami2020trees} with low distortion~\cite{sala2018representation,sarkar2012low}. Since the seminal work of \citet{ganea2018hyperbolic} on Hyperbolic Neural Networks (HNN), approaches  have successfully combined hyperbolic geometry with model architectures ranging from convolutional \citep{shimizu2020hyperbolic} to attention-based \citep{gulcehre2018hyperbolic}, including graph neural networks \citep{liu2019hyperbolic,chami2019hyperbolic} and, most recently, vision transformers \citep{ermolov2022hyperbolic}.
There are two leading interpretations of the hyperbolic radius in hyperbolic space: as a measure of the prediction uncertainty~\citep{chen2022hyperbolic, ermolov2022hyperbolic, franco23, flaborea23} or as the hierarchical parent-to-child relation \citep{nickel2017poincare,tifrea2018poincar,suris2021hyperfuture,ermolov2022hyperbolic,atigh2022hyperbolic}. Our work builds on the SOTA hyperbolic semantic segmentation method of \citet{atigh2022hyperbolic}, which enforces hierarchical labels and training objectives. However, when training without manually injected hierarchical labels, as we do, the hierarchical interpretation does not apply.  
Although a correlation between the hyperbolic radius and an uncertainty measure has been noted, a comprehensive understanding of this relationship is still lacking. In order to further research in this direction, we provide an investigation that examines the relationship between the hyperbolic radius, data scarcity, and epistemic uncertainty, aiming to shed light on this association.
Furthermore, HALO's acquisition score is tailored for semantic segmentation, as it computes the hyperbolic radius for each pixel embedding. Hyperbolic neural networks have shown comparable performance to Euclidean models in semantic segmentation \citep{atigh2022hyperbolic}, enabling fair comparisons. However, this equivalence does not extend to other tasks, where hyperbolic neural networks have not achieved similar performance (image classification) or are yet to be developed (object detection).

\textbf{Active Learning (AL)} \quad
The number of annotations required for dense tasks such as semantic segmentation can be costly and time-consuming. Active learning balances the labeling efforts and performance, selecting the most informative pixels in successive learning rounds.
Strategies for active learning are based on uncertainty sampling \citep{gal2017deep,wang2014new,wang2016cost}, diversity sampling \citep{ash2019deep,kirsch2019batchbald,sener2017active,wu2021redal} or a combination of both \citep{sinha2019variational,xie2022active,prabhu2021active, xie2022ripu}.
For the case of AL in semantic segmentation, EqualAL~\citep{equal2020} incorporates the self-supervisory signal of self-consistency to mitigate the overfitting of scenarios with limited labeled training data. Labor~\citep{shin2021labor} selects the most representative pixels within the generation of an inconsistency mask. PixelPick~\citep{shin2021pixelpick} prioritizes the identification of specific pixels or regions over labeling the entire image. \citet{mittal2023best} explores the effect of data distribution, semi-supervised learning, and labeling budgets. We are the first to leverage the hyperbolic radius as a proxy for the most informative pixels to label next.

\textbf{Active Domain Adaptation (ADA)} \quad
Domain Adaptation (DA) involves learning from a source data distribution and transferring that knowledge to a target dataset with a different distribution.
Recent advancements in DA for semantic segmentation have utilized unsupervised (UDA) \citep{hoffman2018cycada,Vu_2019_CVPR,Yang_2020_CVPR,liu2020open,mei2020instance,liu2021source} and semi-supervised (SSDA) \citep{french2017self,Saito_2019_ICCV,singh2021clda,jiang2020bidirectional} learning techniques. However, challenges such as noise and label bias still pose limitations on the performance of DA methods. 
Active Domain Adaptation (ADA) aims to reduce the disparity between source and target domains by actively selecting informative data points from the target domain \citep{su2020active, fu2021transferable, sing2021, shin2021labor}, which are subsequently labeled by human annotators. In semantic segmentation, \citet{ning2021multi} propose a multi-anchor strategy to mitigate the distortion between the source and target distributions. 
The recent study of \citet{xie2022ripu} shows the advantages of region-based selection in terms of region impurity and prediction uncertainty scores, compared to pixel-based approaches.
By contrast, we show that selecting pixels just from class boundaries limits performance, as they are not necessarily the most informative, as we confirm with an oracular study. Instead, we show that the hyperbolic radius, in conjunction with prediction entropy, effectively approximates epistemic uncertainty, thereby serving as a successful objective for label acquisition.

\textbf{Uncertainty} \quad
The notion of uncertainty has gained increasing attention in machine learning (ML) research in recent years, primarily due to its growing practical significance in real-world applications. Consequently, numerous studies have developed approaches for uncertainty quantification in ML \citep{kendall2017, carvalho20, xiao19, michelmore20}.
Within the existing literature, two distinct sources of uncertainty are commonly acknowledged: aleatoric and epistemic \citep{Fisher1930, Hora1996}. Aleatoric uncertainty stems from the inherent randomness and variability within the data, while epistemic uncertainty arises from a lack of knowledge or data. As a result, epistemic uncertainty can theoretically be reduced with supplementary information, while aleatoric uncertainty remains non-reducible.
Several methodologies have proposed techniques for quantifying both aleatoric and epistemic uncertainty \citep{Depeweg2017, kendall2017, hullermeier21uncert, toro22}. Following the approach of \citet{Depeweg2017}, both total uncertainty and aleatoric uncertainty can be approximated via model ensemble \citep{Lakshminarayanan2016Ensemble}, deriving epistemic uncertainty as the difference between the two. In our study, for the first time, we distinguish two leading complementary causes for epistemic uncertainty: prediction error and data scarcity.

\section{Background}
\label{sec:preliminaries}

\begin{figure*}[ht]
\centering

\includegraphics[width=.9\linewidth]{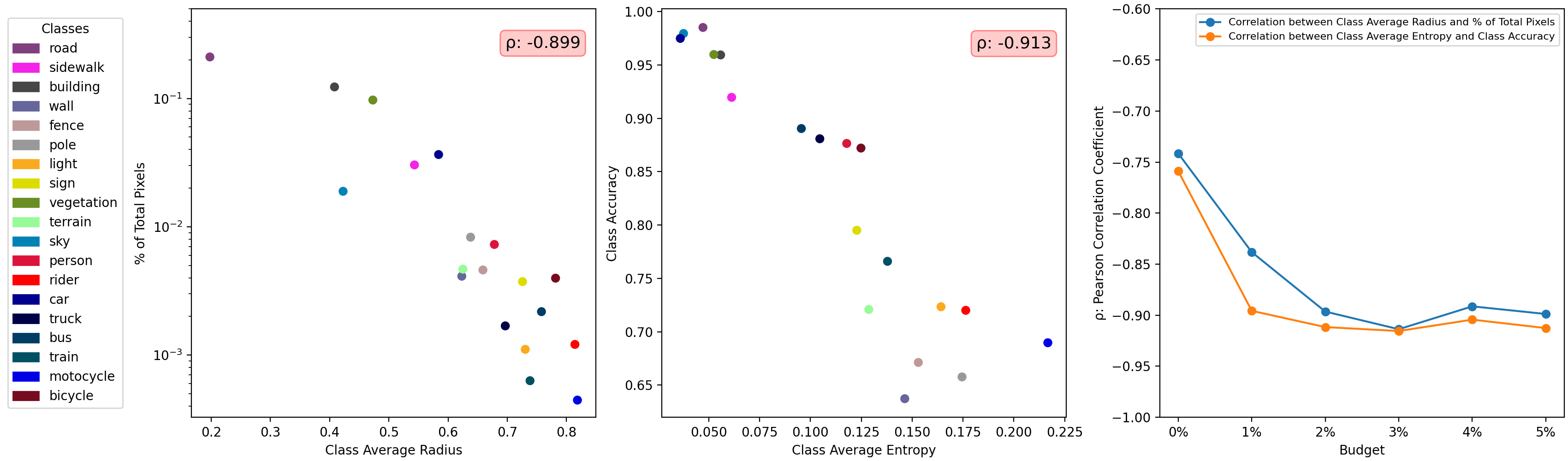}
\caption{(left) Plot of class average radius vs.\ the percentage of total pixels in the target dataset; (center) Plot of the class average entropy vs. class accuracy; (right) Plot of labeling budget vs. correlation between class average radius and percentage of total pixels (blue) and between class average entropy and class accuracy (orange).
}
\label{fig:correlations}

\end{figure*}

In this section, we introduce the background for our work. We begin by discussing Hyperbolic Neural Networks and Hyperbolic Semantic Segmentation, moving then to Active Domain Adaptation, which form the basis of our approach.

\textbf{Hyperbolic Neural Networks and Semantic Segmentation} \quad
We operate in the Poincaré ball hyperbolic manifold.
We define it as the pair $(\mathbb{D}_c^N, g^{\mathbb{D}_c})$ where $\mathbb{D}_c^N = \{x \in \mathbb{R}^N : c \|x\| < 1 \}$ is the manifold and $g_x^{\mathbb{D}_c} = (\lambda_x^c)^2 g^\mathbb{E}$ is the associated Riemannian metric, $-c$ is the curvature, $\lambda_x^c = \frac{2}{1-c\|x\|^2}$ is the conformal factor and $g^\mathbb{E} = \mathbb{I}^N$ is the Euclidean metric tensor.
Hyperbolic neural networks first extract a feature vector $v$ in Euclidean space, which is subsequently projected into the Poincaré ball via exponential map: 
\begin{equation}
    exp_x^c(v) = x \oplus_c \left(\frac{v}{\sqrt{c}\|v\|} tanh\left(\sqrt{c} \frac{\lambda_x^c \|v\|}{2} \right)  \right)
\label{eq:expmap}
\end{equation}
where $x \in \mathbb{D}_c^N$ is the anchor and $\oplus_c$ is the Möbius hyperbolic addition. The latter is defined for two hyperbolic vectors $h, w$ as follows:
\begin{equation}
    h \oplus_c w = \frac{(1 + 2c \langle h,w \rangle +c \|w\|^2)v + (1 - c \|h\|^2)w}{1 + 2c \langle h,w \rangle +c^2 \|h\|^2 \|w\|^2}
\label{eq:mob_add}
\end{equation}
We define the hyperbolic radius of the embedding $h \in \mathbb{D}_c^N $ as the Poincaré distance (See Eq.~\ref{eq:poinc_dist} in Appendix \ref{sec:hyperbolic_formulas}) from the origin of the ball:
\begin{equation}
    d(h, 0) = \frac{2}{\sqrt{c}} tanh^{-1}  \left( \sqrt{c} \| h \|  \right).
\label{eq:hyper_radius}
\end{equation}
For the hyperbolic semantic segmentation, we adopt the work by \citet{atigh2022hyperbolic}, which stands as the first to showcase performance comparable to that of Euclidean networks. Segmentation is performed using hyperbolic multinomial logistic regression (HyperMLR) \citep{ganea2018hyperbolic}. The complete formulation of HyperMLR is in Appendix \ref{sec:hyperbolic_formulas}.

\textbf{ADA for Semantic Segmentation}
\label{sec:ada_ss} \quad
The task aims to transfer knowledge from a source labeled dataset $\mathcal{S}=(X_s, Y_s)$ to a target unlabeled dataset $\mathcal{T}=(X_t, Y_t)$, where $X$ represents an image and $Y$ the corresponding annotation map. $Y_s$ is given, $Y_t$ is initially the empty set $\emptyset$. Adhering to the ADA protocol \citep{xie2022ripu, wu2022d2ada, shin2021labor}, target annotations are incrementally added in rounds, subject to
a predefined budget,  upon querying an annotator. Each pixel is assigned a priority score using a predefined acquisition map $\mathcal{A}$. Labels are added to $Y_t$ in each AL round by selecting pixels from $\mathcal{A}$ with higher scores, in accordance with the budget. Each AL round is divided into two phases. In the first phase, the segmentation model undergoes end-to-end training, with back-propagation incorporating estimates $\hat{Y}_s$ and $\hat{Y}_t$ from the per-pixel cross-entropy loss $\mathcal{L}({\hat{Y}_s,\hat{Y}_t}, {Y_s,Y_t})$. The second phase consists in acquiring new target labels according to the acquisition score $\mathcal{A}$ and the predefined budget.

In Sec. \ref{sec:interpretation}, we assume to have pre-trained the hyperbolic image segmenter of \citet{atigh2022hyperbolic} on the source dataset GTAV~\citep{gtav} and to have domain-adapted it to the target dataset Cityscapes~\citep{cityscapes} through 5 rounds of AL with a total budget of 5\%.
The following analyses consider the radii of the hyperbolic pixel embeddings and the prediction entropy, for which statistics are computed on the Cityscapes validation set.

\section{Hyperbolic Radius and Epistemic Uncertainty} 
\label{sec:interpretation}

In Sec.~\ref{subsec:radius_property} we interpret the emerging properties of hyperbolic radius, and we compare with the interpretations in literature in Sec.~\ref{subsec:additional_insights}. 

\subsection{Emerging properties of the hyperbolic radius}
\label{subsec:radius_property}

\textbf{What does the hyperbolic radius represent?} 
\quad
Fig. \ref{fig:correlations} (left) shows the correlation between the average class hyperbolic radius and the percentage of pixel labels for each class relative to the total number of pixels in the dataset. The correlation is substantial ($\rho=-0.899$), so classes with larger hyperbolic radii such as \emph{motocycle} are rarer in the target dataset, while at lower hyperbolic radii we have more frequent classes such as \emph{road}.
In conclusion, larger hyperbolic radii indicate which classes the model has been exposed less so far in the training.

\textbf{Understanding the role of the prediction entropy} 
\quad
Prior active learning literature~\citep{xie2022active,prabhu2021active, xie2022ripu} agree on the utility of prediction entropy, i.e.\ the entropy of the prediction scores, in the data acquisition strategy.
In Fig.~\ref{fig:correlations} (center) we report the correlation between the class average entropy and the class accuracy, whose resulting value is a strong correlation of $\rho=-0.913$.
In conclusion, prediction entropy appears to be a good indicator for classes with low accuracy. In HALO, we combine prediction entropy with the newly proposed hyperbolic radius.

\textbf{How does learning the hyperbolic manifold proceed?}
\quad
Fig.~\ref{fig:correlations}~(right) illustrates the evolution, across active learning rounds, of the correlation between prediction entropy and accuracy (orange), and the correlation between the hyperbolic radius and the percentage of pixels in the target dataset (blue).
Both correlations exhibit a growing trend in module, eventually saturating at high values.
In conclusion, as the training progresses, both the hyperbolic radius and the prediction entropy become better estimators for data scarcity and prediction error.

\textbf{Relation with the epistemic uncertainty} 
\quad
Following the work of \citet{Depeweg2017}, we quantify the epistemic uncertainty as the difference between total and aleatoric uncertainty. The total uncertainty is estimated by computing the entropy of the predictive posterior distribution $U_t(\bm{x}) = \mathcal{H}[p(y|\bm{x})]$. This formulation encompasses the epistemic uncertainty regarding the network parameters $\bm{\theta}$. To compute it, we first measure the aleatoric uncertainty as $U_a(\bm{x}) = E_{p(\bm{\theta}|\mathcal{D})} \mathcal{H}[p(y|\bm{\theta}, \bm{x})]$ and then we derive the epistemic uncertainty as the difference $U_e(\bm{x}) = U_t(\bm{x}) - U_a(\bm{x})$.
The model ensemble approach \citep{Lakshminarayanan2016Ensemble} offers an effective means to approximate the posterior distribution $p(\bm{\theta}|\mathcal{D})$ using a finite ensemble of models $\bm{\theta}_1, ..., \bm{\theta}_M$. We can approximate the total uncertainty as
\begin{equation}
\scriptsize
    \tilde{U}_t(\bm{x}) = - \sum_{y \in \mathcal{Y}} \left( \frac{1}{M} \sum_{m=1}^M p(y| \bm{\theta}_m, \bm{x}) \right) log_2  \left( \frac{1}{M} \sum_{m=1}^M p(y| \bm{\theta}_m, \bm{x}) \right)
\end{equation}
and similarly the aleatoric uncertainty
\begin{equation}
\small
    \tilde{U}_a(\bm{x}) = -\frac{1}{M} \sum_{m=1}^M \sum_{y \in \mathcal{Y}} p(y| \bm{\theta}_m, \bm{x})\ log_2\ p(y| \bm{\theta}_m, \bm{x}).
\end{equation}
Finally the epistemic uncertainty is approximated by the difference $\tilde{U}_e(\bm{x}) = \tilde{U}_t(\bm{x}) - \tilde{U}_a(\bm{x})$.

The correlation between the epistemic uncertainty and the hyperbolic radius results in a value of $\rho=0.769$, while the correlation between the epistemic uncertainty and the prediction entropy is $\rho=0.789$.
Supported by the fact that the correlation between the hyperbolic radius and the prediction entropy is moderate ($\rho=0.658$), we conclude that they encode complementary signals for the uncertainty description, respectively data scarcity and prediction error. In fact, the correlation between their product and epistemic uncertainty results in an even higher value ($\rho=0.824$). Building upon this observation, we establish the acquisition score as the product of these two metrics (see Sec. \ref{sec:acquisition_strategy}).

\begin{figure}[ht]
\centering
\includegraphics[width=.8\linewidth]{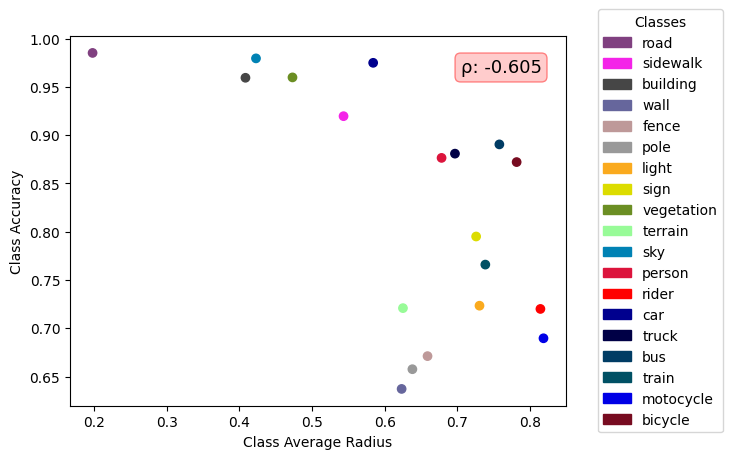}
\vspace{-0.1in}
\caption{Plot of the class average radius vs. class accuracy.}
\vspace{-0.2in}
\label{fig:corr_evolution}
\end{figure}

\subsection{Comparing interpretations of the hyperbolic radius}
\label{subsec:additional_insights}

It emerges from our analysis that larger radii are assigned to classes that have higher data scarcity.
Earlier work has explained the hyperbolic radius in terms of uncertainty or hierarchies. Approaches from the former \citep{franco23, flaborea23} suggest that larger hyperbolic radii indicate more certain and unambiguous samples in terms of classification accuracy. 
In our case, the correlation between the hyperbolic radius and class accuracy, as depicted in Fig. \ref{fig:corr_evolution}, is moderate ($\rho=-0.605$). However, this value is considerably lower than the correlation between the hyperbolic radius and the percentage of pixels. Hence, the radius serves as a more effective indicator of data scarcity (see appendix \ref{sec:selection_priority} for additional analysis).
Another difference with the studies in favor of the uncertainty interpretation lies in the definition of the uncertainty as $1 - \text{radius}$, typical of hyperbolic metric learning-based approaches \citep{franco23, flaborea23}. In those, a larger radius leads to an exponentially larger matching penalty due to the employed Poincaré distance, effectively making the radius inversely proportional to the errors, as those studies show.

Elsewhere, the interpretation of the hyperbolic radius aligns with a hierarchical explanation~\citep{nickel2017poincare,tifrea2018poincar,suris2021hyperfuture,ermolov2022hyperbolic,atigh2022hyperbolic}. These methods involve hierarchical datasets, hierarchical labeling, and classification objective functions. Hierarchies naturally align with the growing volume in the Poincaré ball, resulting in children nodes from different parents being mapped further from each other than from their parents.
Learning under hierarchical constraints results in leaf classes closer to the edge of the ball, and transitions between them traverse their parent nodes at lower hyperbolic radii. 
Our hyperbolic segmentation approach differs from prior hyperbolic works \citep{atigh2022hyperbolic, franco23, flaborea23, ermolov2022hyperbolic} as we employ hyperbolic multinomial logistic regression without the incorporation of hierarchical labels or losses based on the Poincaré distance.
These differences drive our intuition to utilize the hyperbolic radius as an estimator of data scarcity, thereby incorporating it into the final data acquisition score in the active learning process.

\begin{figure*}[ht]
\centering
\includegraphics[trim={0 0 0 0}, scale=0.29]{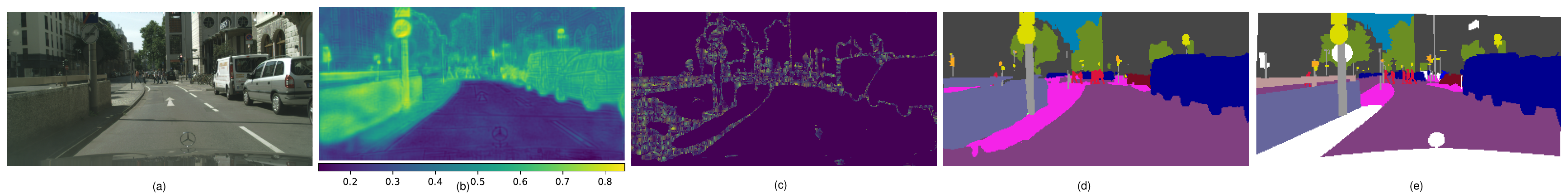}
\caption{(a) Original image;  (b) Radius map depicting the hyperbolic radii of pixel embeddings; (c) Pixels (yellow) that have been selected for data acquisition. See Sec.~\ref{sec:method} for details; (d) HALO prediction; (e) Ground Truth annotations. Zoom in for the details.}
\label{fig:rad_quant}
\end{figure*}


\section{Hyperbolic Active Learning Optimization (HALO)}
\label{sec:method}

In this section, first we introduce the proposed HALO pipeline (Sec.~\ref{sec:halo_pipeline}), then we detail the novel acquisition strategy (Sec.~\ref{sec:acquisition_strategy}). Finally, we present our proposition for fixing the hyperbolic training instability (Sec.~\ref{sec:hfr}).

\subsection{HALO pipeline}
\label{sec:halo_pipeline}

Let us consider Fig.~\ref{fig:teaser}. Our AL strategy consists in assigning an acquisition score $\mathcal{A}$ to each pixel, based on the combination of hyperbolic radius and prediction entropy.
We estimate the hyperbolic radius $\mathcal{R}$ from pixel embeddings (as detailed in Sec.~\ref{sec:interpretation} and illustrated in Fig.~\ref{fig:rad_quant}b). Concurrently, predicted classification probabilities are used to compute the prediction entropy $\mathcal{H}$, inspired by prior works \citep{paul2020domain, shin2021pixelpick, wang2014new, wang2016cost, xie2022ripu}. 
New labels are subsequently chosen based on the acquisition score $\mathcal{A}$ and integrated into the training set (see selected pixels in Fig.~\ref{fig:rad_quant}c).
Note that the new labels are both at the boundaries and within, in areas with the largest inaccuracies (compare Fig.~\ref{fig:rad_quant}d and \ref{fig:rad_quant}e). 

\subsection{Novel data acquisition strategy}
\label{sec:acquisition_strategy}

The acquisition score of each pixel in an image is formulated as the element-wise multiplication of the hyperbolic radius map $\mathcal{R}$ and the prediction entropy map $\mathcal{H}$, i.e.\ $\mathcal{A} = \mathcal{R} \odot \mathcal{H}$. The radius is computed as the distance of the hyperbolic pixel embedding $(i,j)$ from the center of the Poincaré ball $\mathcal{R}^{(i,j)} = d(h_{i,j}, 0)$ (see Eq.~\ref{eq:hyper_radius}).
The prediction entropy $\mathcal{H}^{(i,j)} = -\sum_{c=1}^C P_{i,j,c} \ log P_{i,j,c}$ is estimated as the entropy of the softmax probability array $P_{i,j,c}$ associated with the pixel $(i,j)$ and the classes $c \in \{1,...,C\}$.
The acquisition score $\mathcal{A}$ serves as a surrogate indicator for the epistemic uncertainty of each pixel and determines which ones are presented to the human annotator for labeling, to augment the target label set $Y_t$.

\subsection{Robust hyperbolic learning with feature reweighting}
\label{sec:hfr}

HNNs are prone to instability during training because of the unique topology of the Poincaré ball. More precisely, when embeddings approach the boundary, the occurrence of vanishing gradients can impede the learning process. Several solutions have been proposed in the literature to address this problem. \citet{guo22} achieves robustness by clipping the largest values of the radii, \citet{franco23} makes it by curriculum learning, and \citet{vanspengler23} needs to carefully initialize the hyperbolic network parameters. However, these approaches yield sub-optimal performances or are not compatible with our use case (see Appendix \ref{sec:other_ablations}).
Therefore, we introduce the \textit{Hyperbolic Feature Reweighting (HFR)} module, designed to enhance training stability by reweighting features, prior to their projection onto the Poincaré ball. Given the feature map $Z \in \mathbb{R}^{\tilde{H} \times \tilde{W}}$ generated by the encoder, we compute the weights $L = \textrm{HFR}(Z) \in \mathbb{R}^{\tilde{H} \times \tilde{W}}$ and use them to rescale each entry of the normalized feature map, yielding $\tilde{Z} = \frac{Z}{|Z|} \odot L$, where $|Z| = \sum_{k=1}^{\tilde{H}\tilde{W}} z_{ij}$ and $\odot$ denotes the element-wise multiplication. 
Intuitively, reweighting prevents embeddings from getting too close to the boundaries, where the distances tend to infinity. Our proposed HFR module is end-to-end trained and it enables the model to dynamically adapt through the various stages of training, improving its robustness.

\section{Results}
\label{sec:results}

\begin{table*}[t]
\caption{Comparison of mIoU results for different methods on the (a) \textbf{GTAV$\rightarrow$Cityscapes}, (b) \textbf{SYNTHIA$\rightarrow$Cityscapes}, and (C) \textbf{Cityscapes$\rightarrow$ACDC} benchmarks. Methods marked with $^\sharp$ are based on DeepLab-v3+ \citep{chen2018encoderdecoder}, the ones with $^\dagger$ on SegFormer-B4 \citep{xie2021segformer}, whereas all the others use DeepLab-v2 \citep{deeplab2018}.}
\label{tab:results_table}
\centering
\begin{small}
\resizebox{\textwidth}{!}{
\begin{tabular}{lcccccccccccccccccccccc}
\toprule
Method & Budget & \rot{road} & \rot{side.} & \rot{buil.} & \rot{wall} & \rot{fence} & \rot{pole} & \rot{light} & \rot{sign} & \rot{veg.} & \rot{terr.} & \rot{sky} & \rot{pers.} & \rot{rider} & \rot{car} & \rot{truck} & \rot{bus} & \rot{train} & \rot{motor.} & \rot{bike} & mIoU & mIoU* \\
\toprule
\multicolumn{23}{c}{}\\
\multicolumn{23}{c}{\large (a) GTAV $\rightarrow$ Cityscapes}\\
\midrule
LabOR \citep{shin2021labor} & 2.2\% & 96.6 & 77.0 & 89.6 & 47.8 & 50.7 & \textbf{48.0} & \textbf{56.6} & 63.5 & 89.5 & 57.8 & 91.6 & 72.0 & 47.3 & 91.7 & 62.1 & 61.9 & 48.9 & 47.9 & 65.3 & 66.6 & - \\
RIPU \citep{xie2022ripu} & 2.2\% & 96.5 & 74.1 & 89.7 & 53.1 & 51.0 & 43.8 & 53.4 & 62.2 & 90.0 & 57.6 & 92.6 & 73.0 & 53.0 & \textbf{92.8} & 73.8 & \textbf{78.5} & 62.0 & \textbf{55.6} & 70.0 & 69.6 & - \\
\rowcolor[HTML]{ebebeb}
\textbf{HALO} (ours) & 2.2\% & \textbf{97.5} & \textbf{79.9} & \textbf{90.2} & \textbf{55.6} & \textbf{51.5} & 45.3 & 56.2 & \textbf{66.2} & \textbf{90.2} & \textbf{58.6} & \textbf{92.8} & \textbf{73.3} & \textbf{53.5} & 92.6 & \textbf{76.9} & 76.2 & \textbf{64.2} & 55.2 & \textbf{70.1} & \textbf{70.8} & - \\
\midrule
AADA$^\sharp$ \citep{su2020active} & 5\% & 92.2 & 59.9 & 87.3 & 36.4 & 45.7 & 46.1 & 50.6 & 59.5 & 88.3 & 44.0 & 90.2 & 69.7 & 38.2 & 90.0 & 55.3 & 45.1 & 32.0 & 32.6 & 62.9 & 59.3 & - \\
MADA$^\sharp$ \citep{ning2021multi} & 5\% & 95.1 & 69.8 & 88.5 & 43.3 & 48.7 & 45.7 & 53.3 & 59.2 & 89.1 & 46.7 & 91.5 & 73.9 & 50.1 & 91.2 & 60.6 & 56.9 & 48.4 & 51.6 & 68.7 & 64.9 & - \\
D$^2$ADA$^\sharp$ \citep{wu2022d2ada} & 5\% & 97.0 & 77.8 & 90.0 & 46.0 & 55.0 & 52.7 & 58.7 & 65.8 & 90.4 & 58.9 & 92.1 & 75.7 & 54.4 & 92.3 & 69.0 & 78.0 & 68.5 & 59.1 & 72.3 & 71.3 & - \\
RIPU$^\sharp$ \citep{xie2022ripu} & 5\% & 97.0 & 77.3 & 90.4 & \textbf{54.6} & 53.2 & 47.7 & 55.9 & 64.1 & 90.2 & 59.2 & 93.2 & 75.0 & 54.8 & 92.7 & 73.0 & 79.7 & 68.9 & 55.5 & 70.3 & 71.2 & - \\
\rowcolor[HTML]{ebebeb}
HALO$^\sharp$ (ours) & 5\% & \textbf{97.6} & \textbf{81.0} & \textbf{91.4} & 53.7 & \textbf{54.9} & \textbf{56.7} & \textbf{62.9} & \textbf{72.1} & \textbf{91.4} & \textbf{60.5} & \textbf{94.1} & \textbf{78.0} & \textbf{57.3} & \textbf{94.0} & \textbf{81.4} & \textbf{84.7} & \textbf{70.1} & \textbf{60.0} & \textbf{73.3} & \textbf{74.5} & - \\
\midrule
\rowcolor[HTML]{ebebeb}
\textbf{HALO}$^\dagger$ (ours) & 5\% & \textbf{98.2} & \textbf{85.4} & \textbf{92.5} & \textbf{62.5} & \textbf{61.6} & \textbf{58.3} & \textbf{67.7} & \textbf{74.9} & \textbf{92.2} & \textbf{65.1} & \textbf{94.7} & \textbf{79.9} & \textbf{60.8} & \textbf{94.6} & \textbf{84.1} & \textbf{85.4} & \textbf{83.6} & \textbf{61.2} & \textbf{75.5} & \textbf{77.8} & - \\
\midrule
\midrule
Eucl. Supervised DA$^\sharp$ & 100\% & 97.4 & 77.9 & 91.1 & 54.9 & 53.7 & 51.9 & 57.9 & 64.7 & 91.1 & 57.8 & 93.2 & 74.7 & 54.8 & 93.6 & 76.4 & 79.3 & 67.8 & 55.6 & 71.3 & 71.9 & - \\
Hyper. Supervised DA$^\sharp$ & 100\% & 97.6 & 81.2 & 90.7 & 49.9 & 53.2 & 53.5 & 58.0 & 67.2 & 91.0 & 59.1 & 93.9 & 74.2 & 52.6 & 93.1 & 76.4 & 81.0 & 67.0 & 55.0 & 70.8 & 71.9 & - \\
\midrule

\multicolumn{23}{c}{}\\
\multicolumn{23}{c}{\large (b) SYNTHIA $\rightarrow$ Cityscapes}\\
\midrule
RIPU \citep{xie2022ripu} & 2.2\% & 96.8 & 76.6 & 89.6 & 45.0 & 47.7 & 45.0 & 53.0 & 62.5 & 90.6 & - & \textbf{92.7} & 73.0 & 52.9 & 93.1 & - & 80.5 & - & 52.4 & 70.1 & 70.1 & 75.7 \\
\rowcolor[HTML]{ebebeb}
\textbf{HALO} (ours) & 2.2\% & \textbf{97.5} & \textbf{81.7} & \textbf{90.5} & \textbf{52.8} & \textbf{52.8} & \textbf{45.6} & \textbf{57.3} & \textbf{67.1} & \textbf{91.2} & - & 92.6 & \textbf{74.5} & \textbf{54.9} & \textbf{93.3} & - & \textbf{81.6} & - & \textbf{55.2} & \textbf{71.1} & \textbf{72.5} & \textbf{77.6} \\
\midrule
AADA$^\sharp$ \citep{su2020active} & 5\% & 91.3 & 57.6 & 86.9 & 37.6 & 48.3 & 45.0 & 50.4 & 58.5 & 88.2 & - & 90.3 & 69.4 & 37.9 & 89.9 & - & 44.5 & - & 32.8 & 62.5 & 61.9 & 66.2 \\
MADA$^\sharp$ \citep{ning2021multi} & 5\% & 96.5 & 74.6 & 88.8 & 45.9 & 43.8 & 46.7 & 52.4 & 60.5 & 89.7 & - & 92.2 & 74.1 & 51.2 & 90.9 & - & 60.3 & - & 52.4 & 69.4 & 68.1 & 73.3 \\
D$^2$ADA$^\sharp$ \citep{wu2022d2ada} & 5\% & 96.7 & 76.8 & 90.3 & 48.7 & 51.1 & 54.2 & 58.3 & 68.0 & 90.4 & - & 93.4 & 77.4 & 56.4 & 92.5 & - & 77.5 & - & 58.9 & 73.3 & 72.7 & 77.7 \\
RIPU$^\sharp$ \citep{xie2022ripu} & 5\% & 97.0 & 78.9 & 89.9 & 47.2 & 50.7 & 48.5 & 55.2 & 63.9 & 91.1 & - & 93.0 & 74.4 & 54.1 & 92.9 & - & 79.9 & - & 55.3 & 71.0 & 71.4 & 76.7 \\
\rowcolor[HTML]{ebebeb}
\textbf{HALO}$^\sharp$ (ours) & 5\% & \textbf{97.5} & \textbf{81.5} & \textbf{91.5} & \textbf{56.5} & \textbf{52.7} & \textbf{57.0} & \textbf{63.2} & \textbf{72.9} & \textbf{92.0} & - & \textbf{94.4} & \textbf{77.8} & \textbf{57.4} & \textbf{94.4} & - & \textbf{86.1} & - & \textbf{60.5} & \textbf{73.5} & \textbf{75.6} & \textbf{80.2} \\
\midrule
\rowcolor[HTML]{ebebeb}
\textbf{HALO}$^\dagger$ (ours) & 5\% & \textbf{98.3} & \textbf{86.5} & \textbf{92.6} & \textbf{61.0} & \textbf{61.5} & \textbf{60.6} & \textbf{67.6} & \textbf{76.2} & \textbf{93.2} & - & \textbf{94.6} & \textbf{80.8} & \textbf{58.9} & \textbf{95.0} & - & 85.1 & - & \textbf{62.7} & \textbf{75.6} & \textbf{78.1} & \textbf{82.1} \\
\midrule
\midrule
Eucl. Supervised DA$^\sharp$ & 100\% & 97.5 & 81.4 & 90.9 & 48.5 & 51.3 & 53.6 & 59.4 & 68.1 & 91.7 & - & 93.4 & 75.6 & 51.9 & 93.2 & - & 75.6 & - & 52.0 & 71.2 & 72.2 & 77.1 \\
Hyper. Supervised DA$^\sharp$ & 100\% & 97.7 & 82.2 & 90.3 & 53.0 & 48.8 & 51.7 & 56.0 & 66.1 & 91.4 & - & 94.2 & 75.0 & 51.5 & 93.4 & - & 82.1 & - & 52.8 & 70.2 & 72.3 & 77.1 \\
\midrule

\multicolumn{23}{c}{}\\
\multicolumn{23}{c}{\large (c) Cityscapes $\rightarrow$ ACDC}\\
\midrule
RIPU \citep{xie2022ripu} & 2.2\% & 91.4 & 69.5 & 83.8 & \textbf{52.7} & 41.6 & 52.8 & 66.4 & 54.2 & \textbf{85.1} & 47.5 & \textbf{94.7} & 54.5 & 21.8 & \textbf{85.5} & \textbf{58.7} & 58.8 & \textbf{76.9} & 41.4 & \textbf{45.9} & 62.3 & - \\
\rowcolor[HTML]{ebebeb}
\textbf{HALO} & 2.2\% & \textbf{92.6} & \textbf{71.3} & \textbf{84.5} & 51.3 & \textbf{43.1} & \textbf{53.5} & \textbf{67.2} & \textbf{57.6} & 85.1 & \textbf{49.5} & 94.5 & \textbf{57.2} & \textbf{28.6} & 84.1 & 53.3 & \textbf{76.0} & 66.9 & \textbf{44.1} & 41.4 & \textbf{63.2} & - \\
\midrule
RIPU$^\sharp$ \citep{xie2022ripu} & 5\% & \textbf{92.7} & \textbf{72.5} & 84.7 & 53.1 & 44.8 & 56.7 & 69.1 & 58.9 & 85.9 & 46.9 & \textbf{95.3} & 57.2 & 24.3 & 84.5 & \textbf{61.4} & 59.4 & 79.0 & 36.9 & 43.6 & 63.5 & - \\
\rowcolor[HTML]{ebebeb}
\textbf{HALO}$^\sharp$ & 5\% & 92.6 & 72.2 & \textbf{84.8} & \textbf{54.9} & \textbf{47.7} & \textbf{59.5} & \textbf{71.5} & \textbf{61.1} & \textbf{86.1} & \textbf{49.5} & 95.2 & \textbf{60.7} & \textbf{30.6} & \textbf{85.8} & 58.4 & \textbf{73.8} & \textbf{82.0} & \textbf{41.6} & \textbf{53.2} & \textbf{66.4} & - \\
\midrule
\rowcolor[HTML]{ebebeb}
\textbf{HALO}$^\dagger$ & 5\% & \textbf{95.2} & \textbf{79.8} & \textbf{88.2} & \textbf{60.2} & \textbf{51.1} & \textbf{64.1} & \textbf{78.2} & \textbf{65.6} & \textbf{87.9} & \textbf{55.7} & \textbf{95.5} & \textbf{66.3} & 20.7 & \textbf{88.9} & \textbf{82.2} & \textbf{89.3} & \textbf{87.9} & \textbf{50.4} & \textbf{59.0} & \textbf{71.9} & - \\
\bottomrule
\end{tabular}
}
\end{small}
\end{table*}

In this section, we describe the benchmarks and we perform a comparative evaluation against the SOTA (Sec. \ref{sec:sota}). We conduct ablation studies on the components of HALO and additional analyses in Sec. \ref{sec:ablation} and \ref{sec:other_correlations}. The implementation follows \citet{xie2022ripu} (details in Appendix~\ref{sec:implementation}).

\textbf{Datasets} \quad For pre-training, we utilize GTAV \citep{gtav} and SYNTHIA \citep{synthia} synthetic datasets, each comprising 24,966 and 9,000 densely annotated images, with 19 and 16 classes, respectively. For ADA training and evaluation, we employ real-world target datasets, specifically \textbf{Cityscapes} (CS) train and val sets or \textbf{ACDC} train and test sets, both categorized into the same 19 classes. CS \citep{cityscapes} consists of 2,975 training samples and 500 validation samples. \textbf{ACDC} \citep{acdc} comprises 4,006 images captured under adverse conditions (fog, nighttime, rain, snow) to maximize the complexity and diversity of the scenes.

 \textbf{Training protocol} \quad The model undergoes a pre-training on either GTAV or SYNTHIA source synthetic datasets. Subsequently, the model is domain adapted using both the source and the target datasets. Our hyperbolic radius-based acquisition method is used to select pixels to be labeled in five evenly spaced rounds during training, with either 2.2\% or 5\% of the total labels. Our model is additionally trained under adverse weather conditions, using CS and ACDC as the source and target datasets, respectively, in line with \citet{hoyer2023mic} and \citet{bruggemann2023refign}.
 The ADA performances in Table \ref{tab:results_table} are also compared with the corresponding supervised domain adaptation baselines (Supervised DA). Supervised DA refers to the process where the adaptation to a target dataset involves using all of its labels (i.e., 100\%) for the whole training, in contrast to active domain adaptation which uses a smaller fraction (e.g., 2.2\% or 5\%) of labels.

\textbf{Evaluation metrics} \quad To assess the effectiveness of the models, the mean Intersection-over-Union (mIoU) metric is computed on the target validation set. For GTAV$\rightarrow$CS and CS$\rightarrow$ACDC, the mIoU is calculated on the shared 19 classes, whereas for SYNTHIA$\rightarrow$CS two mIoU values are reported, one on the 13 common classes (mIoU*) and another on the 16 common classes (mIoU).

\subsection{Comparison with the state-of-the-art}
\label{sec:sota}

In Table \ref{tab:results_table}a, we present the results of our method and the most recent ADA approaches on the GTAV$\rightarrow$CS benchmark. HALO outperforms the current state-of-the-art methods \citep{xie2022ripu, wu2022d2ada} using both 2.2\% (+1.2\% mIoU) and 5\% (+3.3\% mIoU) of labeled pixels, reaching 70.8\% and 74.5\%, respectively. Additionally, our method is the first to surpass the supervised domain adaptation baseline (71.9\%), even by a significant margin (+2.6\%). A thorough analysis on the performance at increasing budgets is provided in Sec. \ref{sec:other_correlations}.
HALO achieves state-of-the-art also in the SYNTHIA$\rightarrow$CS case (cf.\ Table \ref{tab:results_table}b), where it improves by +2.4\% and +4.2\% using 2.2\% and 5\% of labels, reaching performances of 72.5\% and 75.6\%, respectively.
Additionally, we train HALO with the SegFormer-B4 \citep{xie2021segformer} segmenter to demonstrate the adaptability of our approach to different architectures. With SegFormer-B4, HALO improves by +3.3\% in GTAV$\rightarrow$CS and +2.5\% in SYNTHIA$\rightarrow$CS compared to HALO with DeepLab-v3+, using 5\% of labels.
Due to the absence of other ADA studies on CS$\rightarrow$ACDC adaptation, we train RIPU~\citep{xie2022ripu} as a baseline for comparison with our method. HALO improves over RIPU by +2.9\% mIoU with a 5\% budget, reaffirming the effectiveness of our approach on a novel dataset, as shown in Table \ref{tab:results_table}c.

\subsection{Ablation study}
\label{sec:ablation}

\begin{table}
\centering
\begin{small}
\caption{Ablation study conducted with the Hyperbolic DeepLab-v3+ as backbone with $5\%$ budget. Performance of prediction entropy and hyperbolic radius scores in isolation (a and b) and combined (c).}
\label{tab:main-ablation}
\resizebox{.7\linewidth}{!}{%
\begin{tabular}{lc}
\toprule
\textbf{Ablative version} & \textbf{mIoU} \\
\midrule
(a)\, Prediction Entropy only ($\mathcal{H}$) & 63.2 \\
(b)\, Hyperbolic Radius only ($\mathcal{R}$) & 64.1 \\
(c)\, \textbf{HALO ($\mathcal{R}\odot\mathcal{H}$)} & \textbf{74.5} \\
\bottomrule
\end{tabular}%
}
\end{small}
\end{table}

We begin by conducting an oracular study using ground-truth labels, followed by ablation studies on the selection criteria, region- versus pixel-based acquisition scores, and HFR. Additional ablation studies are in Appendix~\ref{sec:other_ablations}.

\textbf{Oracle experiment with ground-truth boundaries} \quad
To test our hypothesis that acquiring labels solely from class boundaries results in performance decline, we conduct an oracular experiment. We replace the pseudo-labels used in RIPU with ground-truth labels, effectively evaluating an AL acquisition strategy based on ground-truth boundary pixels. Although oracular, the experiment yields a performance drop of 1.4 mIoU (69.8 vs. to RIPU's 71.2), motivating the design of a novel acquisition strategy which samples also from non-boundary regions.

\textbf{Selection criteria} \quad
HALO demonstrates a substantial improvement of +10.4\% compared to methods (a) and (b) in Table \ref{tab:main-ablation}. More precisely, utilizing solely either the entropy (a) or the hyperbolic radius (b) as acquisition scores yields comparable performance of 63.2\% and 64.1\%, respectively. When these two metrics are combined, the final performance is notably improved to 74.5\%.

\textbf{Region- vs. Pixel-based criteria} \quad
Unlike region impurity in \citet{xie2022ripu}, the hyperbolic radius is a continuous quantity that can be computed for each pixel. We conduct experiments comparing region- and pixel-based acquisition scores. The results demonstrate a small difference between the two approaches (74.1\% vs. 74.5\%), proving that HALO does not necessitate a region-based formulation. More in Appendix~\ref{sec:other_ablations}.

\textbf{Hyperbolic Feature Reweighting (HFR)} \quad HFR improves training stability and enhances performance in the Hyperbolic model. Although the mIoU improvement is modest (+1.6\%), the main advantage is the training robustness, as the Hyperbolic model otherwise struggles to converge. HFR does not benefit the Euclidean model and instead negatively impacts its performance. More in Appendix~\ref{sec:other_ablations}.



\begin{figure}
\centering
\begin{subfigure}[c]{0.47\linewidth}
    \includegraphics[width=\textwidth]{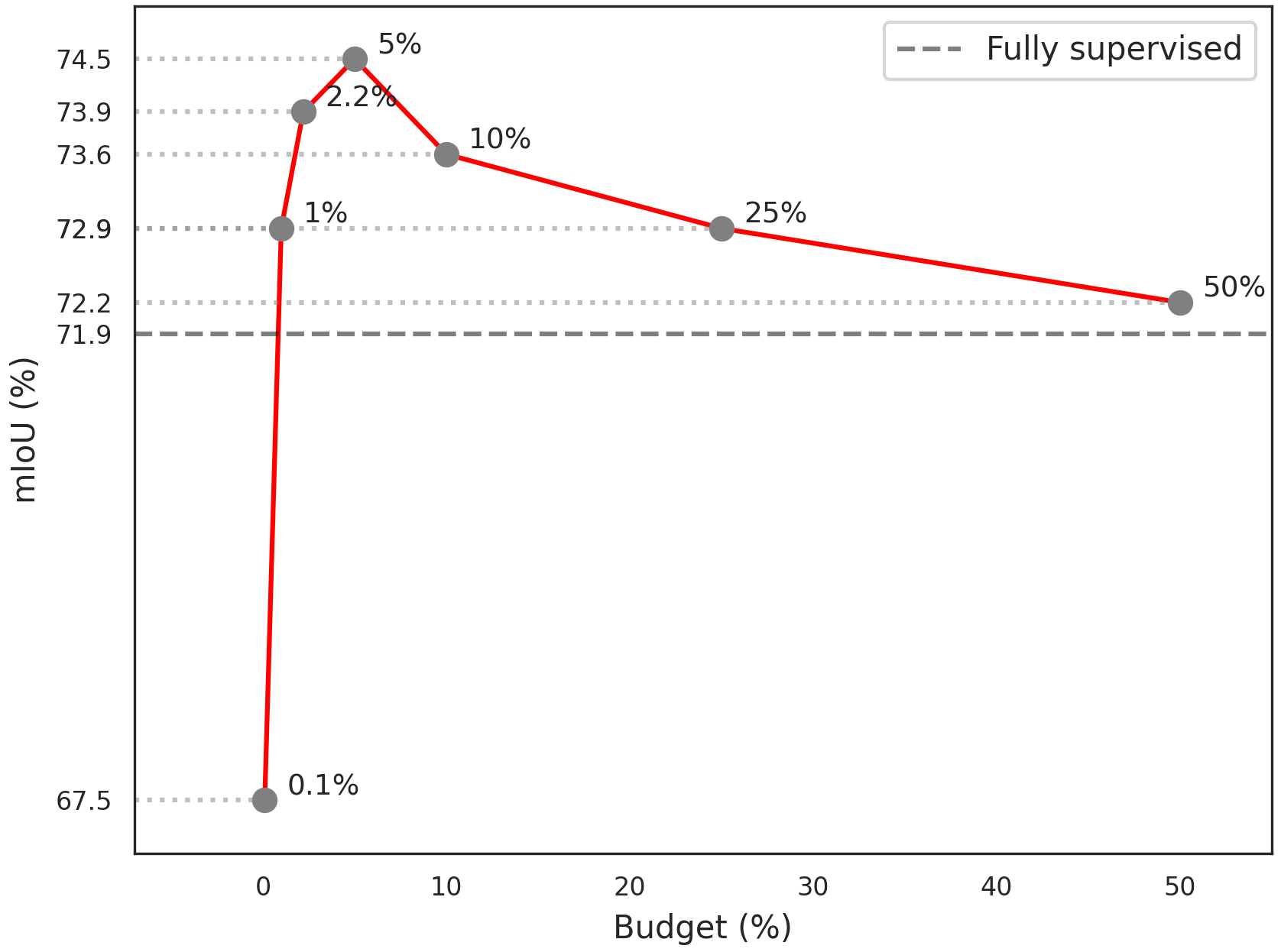}
\end{subfigure}
\hfill
\begin{subfigure}[c]{0.49\linewidth}
    \includegraphics[width=\textwidth]{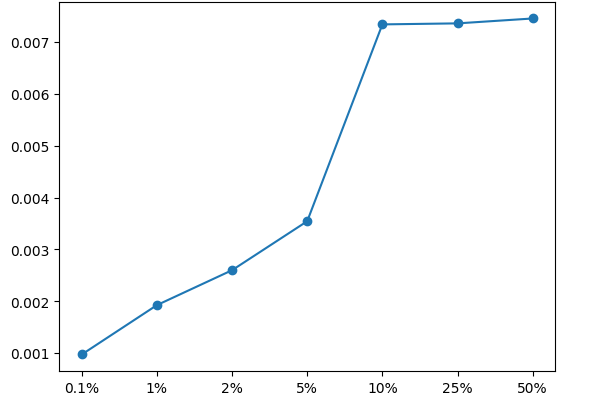}
\end{subfigure}
\caption{(left) Performance on GTAV $\rightarrow$ Cityscapes with different budgets. (right) Evolution of the variance (y axis) of selected pixels distributions with varying budget (x axis).}
\label{fig:budget}
\vspace{-0.2in}
\end{figure}

\subsection{Additional analyses}
\label{sec:other_correlations}

\paragraph{Correlation analysis on Cityscapes$\rightarrow$ACDC} In addition to GTAV$\rightarrow$CS, we report the correlations for CS$\rightarrow$ACDC. The correlation of hyperbolic radius vs. percentage of target pixels is -0.868, while the correlation of prediction entropy vs. class accuracy is -0.892. These results are in line with the GTAV$\rightarrow$CS case (-0.899 and -0.913), showing that the proposed method generalizes well even on a different domain adaptation benchmark.

\textbf{Class imbalance with increasing budget} \quad
We experiment with different labeling budgets, observing performance improvements as the number of labeled pixels increases. However, beyond a threshold of 5\%, adding more labeled pixels leads to diminishing returns (see Fig.~\ref{fig:budget} (left)). We believe this may be explained by data unbalance: taking all labels to domain adapt means that most of them belong to a few classes. Indeed, \textit{road}, \textit{building} and \textit{vegetation} account for 77\% of the total labels, potentially hindering successive training rounds due to data redundancy.

To verify the intuition on data imbalance, we have evaluated the variance of selected pixels distributions as the labelling budget increases.
In Fig.~\ref{fig:budget} (right), we start with the acquisition of just 0.1\% of labels from the target dataset. At this stage, the variance is at a minimum, as HALO manages to identify and select labels from each class in equal proportions. Then the variance increases slowly until the budget reaches 5\%. This happens as the model manages to select pixels from each class, balancing the acquired data selection. The variance has a steep increase at budgets of 10\% and higher. This occurs because the model has already selected most of the labels from the complex and scarce classes which it can identify thanks to the hyperbolic radius and the prediction entropy (cf.\ Sec.~\ref{sec:acquisition_strategy}). So, for budgets of 10\% or more, the data acquisition strategy is influenced by the target dataset imbalance.
The imbalance trend in label selection matches the performance variation in Fig.~\ref{fig:budget} (left). Therefore, we conclude that HALO's selection aids performance, beyond the supervised domain adaptation, until the model manages to successfully identify complex and scarce classes, and until they are available in the target dataset.
\section{Conclusions}
\label{sec:conclusion}
We have introduced the first hyperbolic neural network technique for AL, which we have extensively validated as the novel state-of-the-art on semantic segmentation under domain shift.
We have identified a novel interpretation of the hyperbolic radius as an estimator of data scarcity and epistemic uncertainty, and we have supported the finding with experimental evidence.
The novel concept of hyperbolic radius and its successful use as an acquisition strategy in AL are a step forward in understanding hyperbolic neural networks.
\section*{Limitations}
\label{sec:limitations}

While we have presented experimental evidence supporting the need for a novel interpretation of the hyperbolic radius, our work lacks a rigorous mathematical validation of the properties of the hyperbolic radius within the given experimental setup. Future research should delve into this mathematical aspect to formalize and prove these properties.

HALO's reliance on a source model pretrained on synthetic data introduces challenges related to large-scale simulation efforts and the need for effective synthetic-to-real domain adaptation. Exploring alternative strategies, such as self-supervised pre-training on real source datasets, could be a promising research direction to mitigate these challenges.

Although Active Domain Adaptation significantly reduces labeling costs, the manual annotation of individual pixels can be a time-consuming task. Further investigation into human-robot interaction methodologies to streamline pixel annotation processes and expedite the annotation workflow is needed.
\section*{Acknowledgements}
\label{sec:ack}

We gratefully acknowledge Panasonic Corporation for funding this study.
We also acknowledge financial support from the PNRR MUR project PE0000013-FAIR and from Regione Lazio (Italy), PO FSE 2014-2020 program.
\section*{Impact Statement}
\label{sec:impact}

Hyperbolic Neural Networks (HNN) have recently gained prominence, achieving state-of-the-art performance across various tasks. However, the theory and interpretation of HNN remain diverse, particularly concerning the interpretation of the hyperbolic radius. While it has been traditionally viewed as a continuum hierarchical parent-to-child measure or as an estimate of uncertainty, our novel interpretation adds a new dimension to the growing framework of HNN, advancing the field further.

While there are many potential societal consequences of our work, we feel none must be specifically highlighted here.

\bibliography{references}
\bibliographystyle{icml2024}

\newpage
\appendix
\onecolumn

\setcounter{section}{0}
\renewcommand{\thesection}{A.\arabic{section}}
\setcounter{equation}{0}
\renewcommand{\theequation}{A\arabic{equation}}
\setcounter{table}{0}
\renewcommand{\thetable}{A\arabic{table}}

\section*{Appendix}
\label{sec:appendix}

This appendix provides additional information and insights on the proposed Hyperbolic Active Learning Optimization (HALO) for semantic segmentation under domain shift.

This supplementary material is structured as follows:

\begin{description}
    \item [\ref{sec:other_ablations}: Additional ablation studies] presents additional ablation studies on the proposed Hyperbolic Feature Reweighting (HFR), strategies for stable training in hyperbolic space, region- vs. pixel-based acquisition score, evaluation on the source-free protocol and correlation between Riemannian variance and classification accuracy;
    \item [\ref{sec:hyperbolic_formulas} Additional hyperbolic formulas] reports additional employed hyperbolic formulas;
    \item [\ref{sec:implementation} Implementation details] describes the training details adopted in the experiments;
    \item [\ref{sec:params} Comparison of parameters count] compares the number of parameters of HALO with the baseline model;
    \item [\ref{sec:resources} Computational resources consumption] analyzes the computational cost of HALO compared to the baseline;
    \item [\ref{sec:qualitative} Qualitative results] showcases representative qualitative results of HALO;
    \item [\ref{sec:selection_priority} Data acquisition strategy: rounds of selection] illustrates examples of pixel labeling selection and the priorities of the data acquisition strategy at each acquisition round;
    \item [\ref{sec:additional_comparison} Qualitative comparison with the baseline model] illustrates a qualitative comparison of pixel acquisition between HALO and baseline model to prove the limitation of boundary-only selection;  
\end{description}

\section{Additional ablation studies}
\label{sec:other_ablations}

\subsection{Results of HFR}
Table \ref{tab:hfr} provides insights into the performance of hyperbolic and Euclidean models with and without Hyperbolic Feature Reweighting (HFR). In the case of HALO, the performance with and without HFR remains the same in the source-only setting. However, when applied to the source+target ADA scenario, HFR leads to an improvement of 1.2\%. It should be noted that HFR also stabilizes the training of hyperbolic models. In fact, when not using HFR, training requires a warm-up schedule and, still, it does not converge in approximately 20\% of the runs.
HFR improves therefore performance for ADA and it is important for hyperbolic learning stability.
\begin{table}[h]
\centering
\caption{\textbf{HFR Performance Comparison:} Evaluating the impact of Hyperbolic Feature Reweighting (HFR) on hyperbolic and Euclidean models in source-only and source+target protocols.}
\label{tab:hfr}
\begin{tabular}{lccc}
\toprule
Encoder & Protocol & HFR & mIoU (\%) \\
\midrule
DeepLab-v3+ & source-only & \ding{55} & 36.3 \\
DeepLab-v3+ & source-only & \checkmark & 22.7 \\
Hyper DeepLab-v3+ & source-only & \ding{55} & 39.0 \\
Hyper DeepLab-v3+ & source-only & \checkmark & 38.9 \\
\midrule
HALO & source$+$target & \ding{55} & 72.9 \\
\rowcolor[HTML]{ebebeb}
\textbf{HALO} & \textbf{source$+$target} & \checkmark & \textbf{74.5} \\
\bottomrule
\end{tabular}
\end{table}

\subsection{Exploring strategies for stable training in hyperbolic space}
Here we conduct a more comprehensive evaluation of approaches aimed at stabilizing the training of hyperbolic neural networks. We test \citet{guo2022clipped}'s Feature Clipping method in our framework for comparison with our HFR. As shown in the Table \ref{tab:stabilization}, while Feature Clipping works and produces better results than the baseline RIPU, it still falls short of our HFR method (-1.2\%). \citet{guo2022clipped} utilize Feature Clipping to prevent vanishing gradients during backpropagation. Despite its simplicity, this technique restricts the model's representational capacity by clipping features, resulting in inferior performance compared to our HFR. We have not tested the curriculum learning of \citet{franco23} and the initialization approach of \citet{vanspengler23}, because their adaptation to the ADA task is not straightforward.
The curriculum learning in \citet{franco23} is specifically tailored for metric learning scenarios involving a hyperbolic loss, enabling training in hyperbolic space by utilizing cosine distance for improved initialization, gradually transitioning to the Poincaré loss. Our method does not involve comparing embeddings and leveraging the Poincaré loss. Similarly, the initialization approach in \citet{vanspengler23} is designed explicitly for fully hyperbolic ResNets, particularly hyperbolic convolutions. As we do not employ hyperbolic convolutional layers, their initialization approach is not immediately suitable for our model.
\begin{table}[h]
\centering
\caption{Comparison of strategies for stable training in hyperbolic space}
\label{tab:stabilization}
\begin{tabular}{lc}
\toprule
Method & mIoU (\%) \\
\midrule
\rowcolor[HTML]{ebebeb}
\textbf{Hyperbolic Feature Reweighting} (ours) & \textbf{74.5} \\
Feature Clipping (Guo et al., 2022) & 73.3 \\
Initialization (van Spengler et al., 2023) & not compatible \\
Curriculum Learning (Franco et al., 2023) & not compatible \\
\bottomrule
\end{tabular}
\end{table}

\subsection{Region- vs. Pixel-based acquisition score}
While the region impurity score of RIPU \citep{xie2022ripu} requires pixel regions to work, as the impurity is based on region statistics, the hyperbolic radius employed in HALO can be computed on both pixel and region bases. Here we train HALO with the region-based approach for comparison. As we observe in the Table \ref{tab:region_vs_pixel}, the region-based approach leads to a small difference of -0.4\% on the GTAV$\rightarrow$CS benchmark with 5\% acquired labels, but still manages to achieve a significative improvement over the baseline (RIPU).
\begin{table}[h]
\centering
\caption{Comparison of Region- vs. Pixel-based acquisition score.}
\label{tab:region_vs_pixel}
\begin{tabular}{lccc}
\toprule
Method & Region-based & Pixel-based & mIoU (\%) \\
\midrule
RIPU \citep{xie2022ripu} & \checkmark &  & 71.2 \\
\rowcolor[HTML]{ebebeb}
HALO (ours) & \checkmark &  & 74.1 \\
\rowcolor[HTML]{ebebeb}
\textbf{HALO} (ours) & & \checkmark & \textbf{74.5} \\
\bottomrule
\end{tabular}
\end{table}

\subsection{Source-free domain adaptation}
In the source-free protocol, the model is pre-trained on the source dataset and domain-adapted using only the target dataset. In Table \ref{tab:source-free} we show the performance of HALO on the source-free GTAV $\rightarrow$ CS domain adaptation task. HALO surpasses the current best~\cite{xie2022ripu} by +3\% using 2.2\% of labels.
\begin{table}[h]
\centering
\caption{HALO performance on the \textbf{source-free protocol} on GTAV$\rightarrow$CS, compared with the previous state-of-the-art approach. Methods marked with $^\sharp$ are based on DeepLab-v3+ \citep{chen2018encoderdecoder}, whereas all the others use DeepLab-v2 \citep{deeplab2018}.}
\label{tab:source-free}
\begin{tabular}{lcc}
\toprule
\textbf{Method} & \textbf{Budget} & \textbf{mIoU} \\
\midrule
RIPU \citep{xie2022ripu} & 2.2\% & 67.1 \\
\rowcolor[HTML]{ebebeb}
\textbf{HALO} (ours) & 2.2\% & \textbf{70.1} \\
\midrule
\rowcolor[HTML]{ebebeb}
\textbf{HALO}$^\sharp$ (ours) & 5\% & \textbf{73.3} \\
\bottomrule
\end{tabular}
\end{table}

\subsection{Analysis on the Riemannian variance}
\begin{figure}[h]
\centering
\includegraphics[width=.5\linewidth]{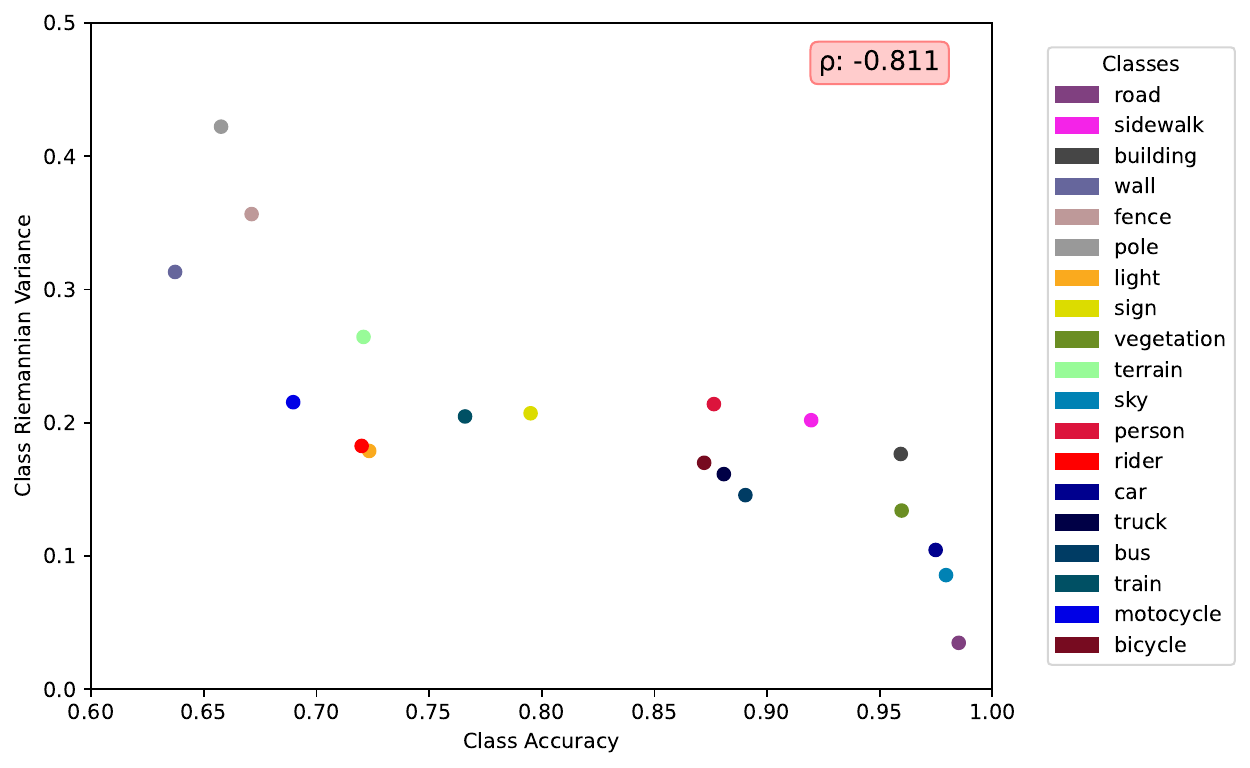}
\caption{Plot of per-class accuracy against per-class Riemannian variance.}
\label{fig:acc_var_corr}
\end{figure}

Fig.~\ref{fig:acc_var_corr} complements our analysis by plotting the class accuracies vs.\ the Riemannian variance (see Eq. \ref{eq:riem_var} in Appendix \ref{sec:hyperbolic_formulas}) of radii for each class. The latter generalizes the Euclidean variance, considering the increasing Poincaré ball density at larger radii. The correlation between accuracy and Riemannian variance is noteworthy ($\rho=-0.811$), indicating that challenging classes, like \emph{pole}, exhibit lower accuracy and larger Riemannian variance, occupying a greater volume in the space.

\section{Additional hyperbolic formulas}
\label{sec:hyperbolic_formulas}

Here we report established hyperbolic formulas and definitions which have used in the paper, but not shown due to space constraints.

\paragraph{Poincaré Distance} Given two hyperbolic vectors x, y $\in \mathbb{D}^N_c$, the \emph{Poincaré distance} represents the distance between them in the Poincaré ball and is defined as:
\begin{equation}
\label{eq:poinc_dist}
    d_{Poin}(x,y) = \frac{2}{\sqrt{c}} \ tanh^{-1} (\sqrt{c} \| -x \oplus_c y\|)
\end{equation}
where $\oplus_c$ is the Möbius addition defined in Eq. 2 of the paper and $c$ is the manifold curvature.

\paragraph{Riemannian Variance} Given a set of hyperbolic vectors $x_1, ..., x_M \in \mathbb{D}^N_c$ we define the Riemannian variance between them as:
\begin{equation}
\label{eq:riem_var}
    \sigma^2 = \frac{1}{M} \sum_{i=1}^M d_{Poin}^2(x_i, \mu)
\end{equation}
where $\mu$ is the Fréchet mean, the hyperbolic vector that minimizes the Riemannian variance. $\mu$ cannot be computed in closed form, but it may be approximated with a recursive algorithm~\cite{lou2021differentiating}.

\paragraph{Hyperbolic Multinomial Logistic Regression (MLR)}
Following~\citet{ganea2018hyperbolic}, to classify an image feature $z_i \in \mathbb{R}^{N}$ we project it onto the Poincaré ball $h_{i} = \exp_x^c(z_{i}) \in \mathbb{D}_c^{N}$ and classify with a number of hyperplanes $H_y^c$ (known as "gyroplanes") for each class $y$:
\begin{equation}
    H_y^c = \{ h_{i} \in \mathbb{D}_c^N, \langle -p_y \oplus_c h_{i}, w_y \rangle \},
\label{eq:gyroplane}
\end{equation}
where, $p_y$ represents the gyroplane offset, and $w_y$ represents the orientation for class $y$.
The distance between a Poincaré ball embedding $h_{i}$ and the gyroplane $H_y^c$ is given by:
\begin{equation}
    d(h_{i}, H_y^c) = \frac{1}{\sqrt{c}} sinh^{-1}  \left( \frac{2 \sqrt{c} \langle -p_y \oplus_c h_{i}, w_y \rangle}{(1-c\|-p_y \oplus_c h_{i}\|^2) \|w_{y}\|}  \right),
\label{eq:hyper_dist}
\end{equation}
Based on this distance, we define the likelihood as $p(\hat{y}_{i} = y | h_{i}) \propto exp(\zeta_y(h_{i}))$ where $\zeta_y(h_{i}) = \lambda_{p_y}^c \|w_{y}\| d(h_{i}, H_y^c)$ is the logit for the $y$ class.

\section{Implementation details}
\label{sec:implementation}

For all experiments, the model is trained on 4 Tesla V100 GPUs using PyTorch \cite{pytorch} and PyTorch Lightning with an effective batch-size of 8 samples (2 per GPU). The DeepLab-v3+ architecture is initialized with an Imagenet pre-trained ResNet-101 as the backbone. \textit{RiemannianSGD} optimizer with momentum of $0.9$ and weight decay of $5 \times 10^{-4}$ is used for all the trainings. The base learning rates for the encoder and decode head are $1 \times 10^{-3}$ and $1 \times 10^{-2}$ respectively, and they are decayed with a "polynomial" schedule with power $0.5$. The models are pre-trained for 15K iterations and adapted for an additional 15K on the target set. As per \cite{xie2022ripu}, the source images are resized to $1280 \times 720$, while the target images are resized to $1280 \times 640$.

\section{Comparison of parameters count}
\label{sec:params}

To provide additional insights into the hyperbolic architecture employed, we conduct a comparison of parameter counts between RIPU \citep{xie2022ripu} and our method HALO (see Table \ref{tab:parameters}). Both employ the DeepLab-v3+ architecture but with some distinctions. RIPU operates with a pixel embedding dimension of 512, resulting in a parameter count of 60.1M. In contrast, HALO operates with a reduced pixel embedding dimension of 64, which the adoption of a hyperbolic learning enables. Moreover, the HyperMLR requires fewer parameters than the Euclidean Linear layer used for classification due to the reduced embedding dimension. This results in a slightly lower total parameter count than RIPU's (10k fewer params). Additionally, HALO introduces the HFR module, consisting of two linear layers separated by a BatchNorm layer and a ReLU. Thanks to the lower embedding dimensions, the input and output sizes of the HFR module are only 64-dimensional, adding less than 10k additional parameters. This roughly matches the number of parameters removed from the segmenter. These modifications result in the parameter count being nearly identical between the two methods (60.1M), aligning with other studies leveraging the DeepLab-v3+ architecture.
\begin{table}[h]
\centering
\caption{Comparison of parameters count in HALO vs. RIPU \citep{xie2022ripu}.}
\label{tab:parameters}
\begin{tabular}{lcccc}
\toprule
Method & Segmenter & Dim. & HFR (params) & Total Params \\
\midrule
RIPU & DeepLab-v3+ & 512 & Not used & 60.1M \\
\rowcolor[HTML]{ebebeb}
HALO (ours) & Hyper-DeepLab-v3+ & 64 & 10k & 60.1M \\
\bottomrule
\end{tabular}
\end{table}

\section{Computational Resources}
\label{sec:resources}

We evaluated the computational resources required by our method, HALO, compared to the previous state-of-the-art, RIPU, using the setup described in Appendix \ref{sec:implementation}. The comparison was conducted under the source+target protocol.

\subsection{Environment and Computational Load Metrics}
Using an identical environment — the same conda environment, 4 Tesla V100 GPUs, and a batch size of 2 per GPU — we compared different computational load metrics for HALO and RIPU using DeepLab-v3+. Table \ref{tab:comp_load_metrics} summarizes the comparison:

\begin{table}[h]
\centering
\caption{Comparison of computational load metrics for HALO and RIPU \citep{xie2022ripu}.}
\label{tab:comp_load_metrics}
\begin{tabular}{lccc}
\toprule
Method & FLOPS $\downarrow$ & FPS $\uparrow$ & Params $\downarrow$ \\
\midrule
RIPU & 125.49 M & 5.62 & 60.1 M \\
\rowcolor[HTML]{ebebeb}
HALO & 280.17 M & 4.72 & 60.1 M \\
\bottomrule
\end{tabular}
\end{table}

FLOPs (Floating Point Operations) are calculated only for the classification layers, which differ between the RIPU and HALO segmentation models. The rest of the model architectures require 1.7 TFLOPS. FPS (Frames Per Second) is measured at inference time. The parameter count (Params) refers to the entire model.

Both models have identical parameter counts and are trained using the same batch size, resulting in negligible differences in memory consumption.

\subsection{Training and inference times}
Training RIPU takes 12.5 hours, while training HALO takes 13.5 hours, which is just an 8\% increase. This marginal difference is primarily due to the nature of operations in hyperbolic space, such as Möbius addition. Despite this, the increased computational cost of hyperbolic operations is offset by the reduced embedding size needed to achieve state-of-the-art performance in hyperbolic space, as detailed in Appendix \ref{sec:params}.

At inference time, evaluated on the Cityscapes validation set, RIPU takes 1m:29s, while HALO takes 1m:46s. Again, the difference in computing time is minor.

\subsection{Optimization considerations}
It is important to note that existing implementations of hyperbolic operations are still under active development and may not yet be fully optimized for mainstream deep learning frameworks and hardware. Specifically, the primary hyperbolic operations — exponential mapping (Eq.~\ref{eq:expmap}), Möbius addition (Eq.~\ref{eq:mob_add}), and Poincaré distance (Eq.~\ref{eq:poinc_dist}) — have not been optimized to the same extent as their Euclidean counterparts, which have benefited from over a decade of optimization.

This analysis demonstrates that the computational overhead introduced by hyperbolic operations is manageable and does not significantly impact the overall efficiency of our method.

\section{Qualitative results}
\label{sec:qualitative}

\begin{figure*}[t]
\centering
\includegraphics[width=\textwidth]{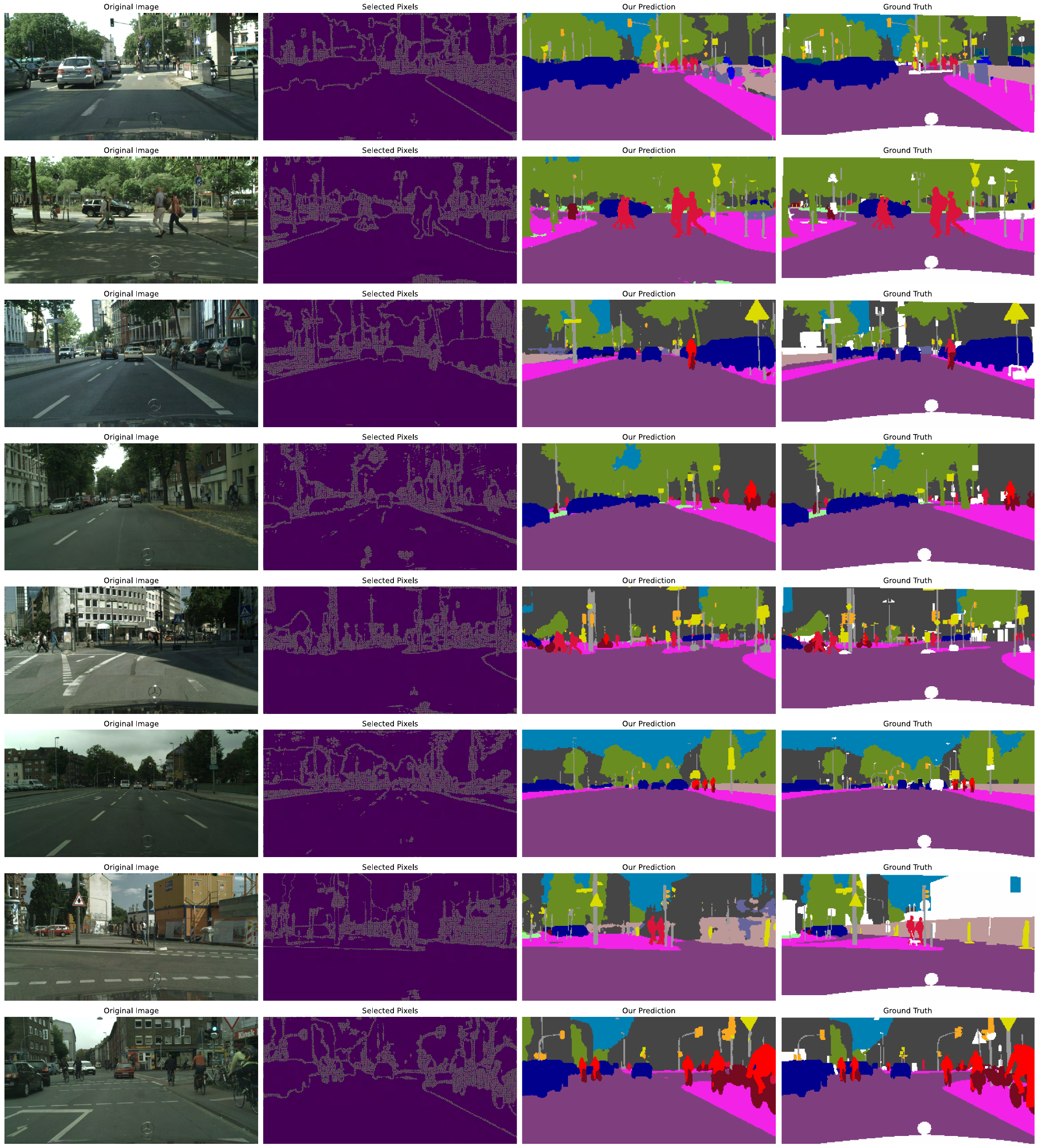}
\caption{\textbf{Qualitative Results Visualization for the GTAV $\rightarrow$ Cityscapes Task.} The figure showcases different subfigures representing: the original image, HALO's pixel selection, HALO's prediction, and the ground-truth label. Zoom in for the  details.}
\label{fig:qualitative_1}
\end{figure*}

In Fig.~\ref{fig:qualitative_1}, we present visualizations of HALO's predicted segmentation maps and the selected pixels.
In the first row, HALO prioritizes the selection of pixels that are not easily interpretable, as evident in the \textit{fence} or \textit{wall} on the right side of the image. Notably, HALO does not limit itself to selecting contours exclusively; it continues to acquire pixels within classes if they exhibit high acquisition score.
This behavior is also observed in rows 2, 3, and 4 of Fig.~\ref{fig:qualitative_1}. For classes with lower complexity, such as \textit{road} and \textit{car}, HALO acquires only the contours. However, for more intricate classes like \textit{pole} and \textit{signs}, it also selects pixels within the class.

In rows 5, 6, and 8, the images depict a crowded scene with numerous small objects from various classes. Remarkably, the selection process directly targets the more complex classes (such as \textit{pole} and \textit{signs}), providing an accurate classification of these.
In row 7, we observe an example where the most common classes (\textit{road}, \textit{vegetation}, \textit{building}, \textit{sky}) dominate the majority of the image. HALO efficiently allocates the labeling budget by focusing on the more complex classes, rather than expending resources on these prevalent ones.
Refer to Sec. \ref{sec:selection_priority} and Fig.~\ref{fig:evolution_selection} for a detailed overview of the selection prioritization during each active learning round.

\section{Data acquisition strategy: rounds of selection}
\label{sec:selection_priority}

\begin{figure*}[t]
\centering
\includegraphics[trim={0 0 0 0}, width=\linewidth]{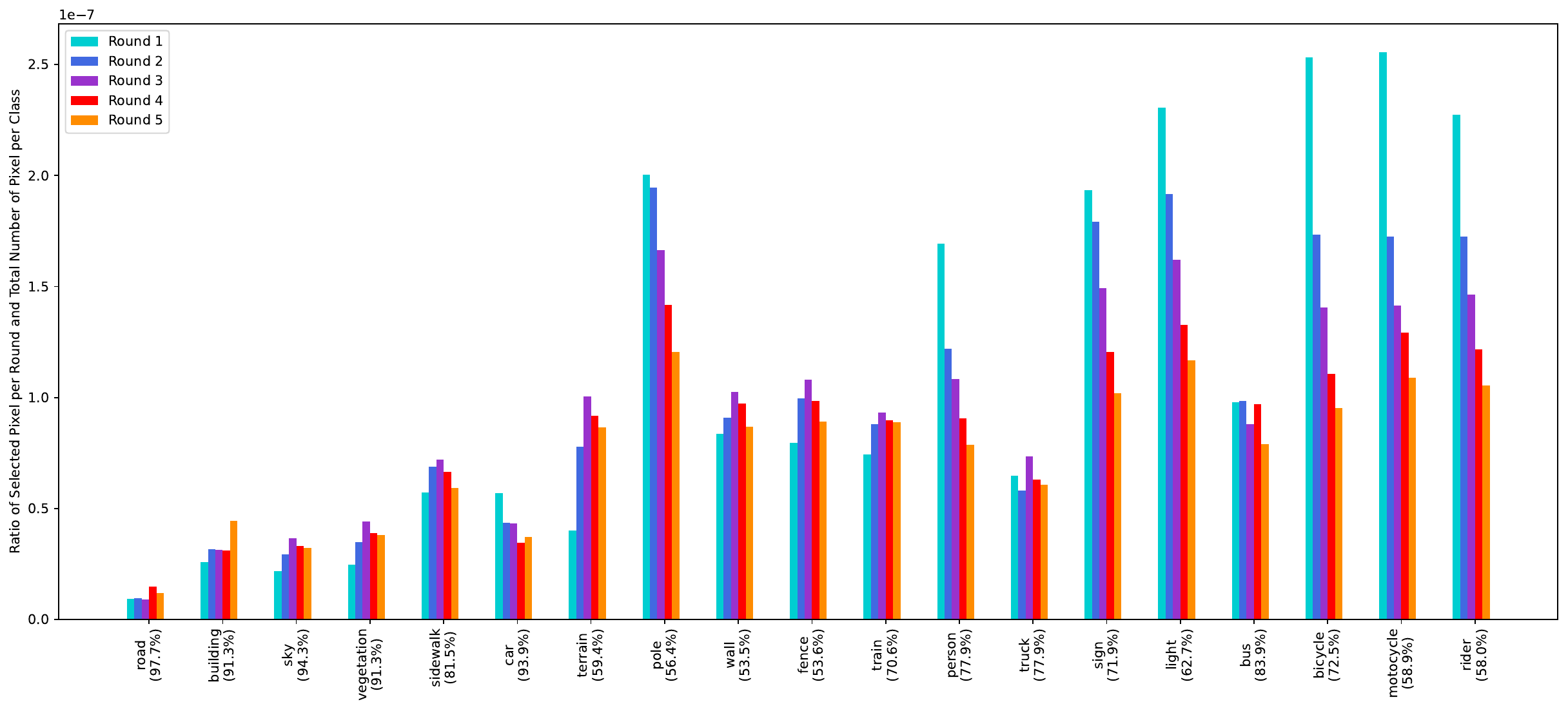}
\caption{Ratio between the selected pixels for each class at each round and the total number of pixels per class. Each color shows the ratio in the specific round. On the $x$-axis are reported the classes with the relative mIoU (\%) of HALO (cf.\ Table 1 of the main paper) ordered according to they decreasing hyperbolic radius.}
\label{fig:selection_distribution}
\end{figure*}

In this section, we analyze how the model prioritizes the selection of the pixels during the different rounds. 
In Fig.~\ref{fig:selection_distribution}, we consider the ratio between the selected pixel at each round and the total number of pixels for the considered class. Note how the model selects in the early stages from the class with high intrinsic difficulty (e.g., rider,  bicycle, pole). During the different rounds, the number of selected pixels decreases because of the scarcity of pixels associated with these classes. On the other hand, less complex classes are less considered in the early stages and the model selects from them in the intermediate rounds if the class has an intermediate complexity (e.g., wall, fence, sidewalk) or in the last stages if it has low complexity (e.g., road or building).

\begin{figure*}[t]
\centering
\includegraphics[width=\textwidth]{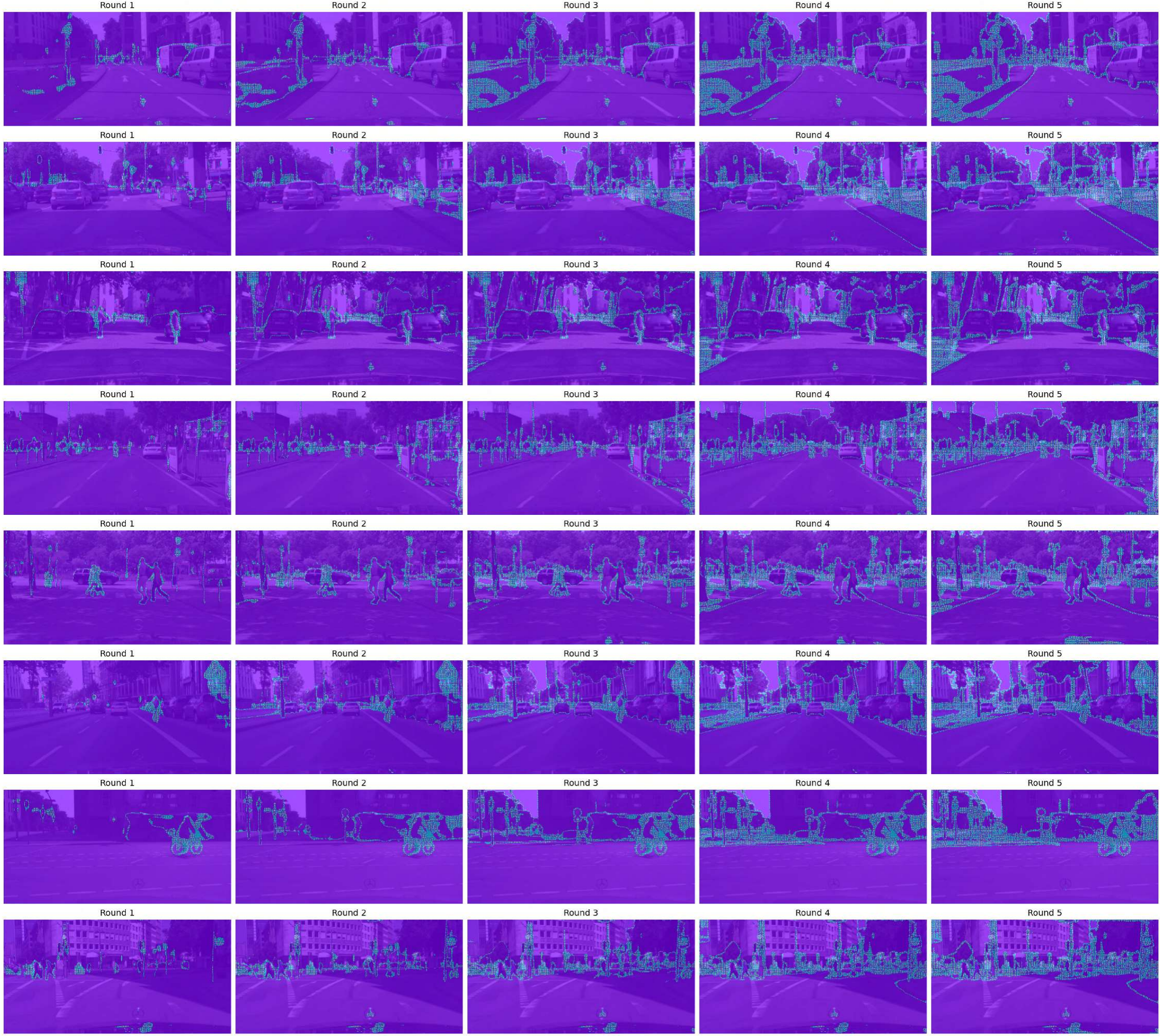}
\caption{Qualitative analysis on the pixel selected by HALO at each round. Zoom in to see details.}
\label{fig:evolution_selection}
\end{figure*}

The qualitative samples of pixel selections in Fig.~\ref{fig:evolution_selection} corroborate this observation. In rounds 1 and 2, the model gives precedence to selecting pixels from more complex classes (e.g., \textit{poles}, \textit{sign}, \textit{person}, or \textit{rider}). Subsequently, HALO shifts its focus to two distinct objectives: i) acquiring contours from classes with lower complexity (e.g., \textit{road}, \textit{car}, or \textit{vegetation}), and ii) obtaining additional pixels from more complex classes (e.g., \textit{pole} or \textit{wall}).
Notably, in rows 1, 2, 3, 5, and 6, HALO gives priority to selecting complete objects right from the initial round (as seen with the \textit{sign}). Another noteworthy instance is the acquisition of the \textit{bicycle} in row 7. The hyperbolic radius score enables the acquisition of contours that extend beyond the boundaries of pseudo-label classes. In this case, we observe precise delineation of the internal portions of the wheels.

\section{Qualitative comparison with the baseline model}
\label{sec:additional_comparison}

\begin{figure*}[ht]
\centering
\includegraphics[width=1.\textwidth]{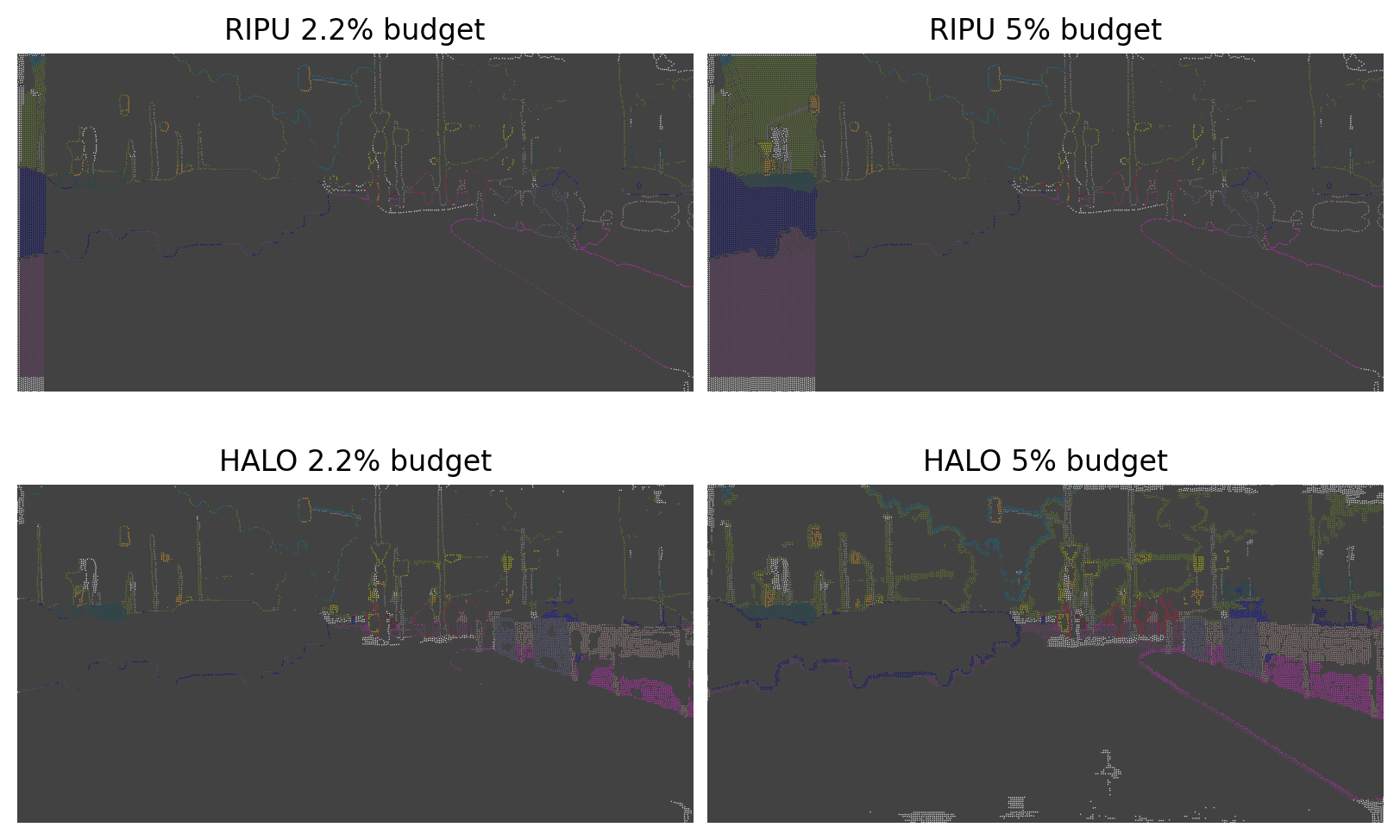}
\caption{(top-row) Pixel selection with RIPU's baseline; (bottom-row) Pixel selection with out HALO; (left-column) Selection with budget 2.2\%; (right-column) Selection with budget 5\%. Zoom in for the details. 
}
\label{fig:active_mask_comparison}
\vspace{-0.2cm}
\end{figure*}

The top row of Fig. \ref{fig:active_mask_comparison} depicts label acquisition using the baseline RIPU method with budgets of 2.2\% (left) and 5\% (right). The bottom row illustrates visualizations with our proposed HALO using the same budgets. Noteworthy observations include:
\begin{itemize}
    \item By design, RIPU only concentrates on selecting boundaries between semantic parts (ref. Fig. \ref{fig:active_mask_comparison} top-left). However, since there are only a few (thin) boundary pixels, RIPU soon exhausts the pixel selection request. Next, when a larger budget is available, RIPU simply samples from the left side. The random selection still provides additional labels (ref. Fig. \ref{fig:active_mask_comparison} top-right) and is a good baseline, cf. Table 3 of ~\citet{xie2022ripu}, although not as good as HALO's acquisition strategy.
    \item By contrast, HALO showcases pixel selection from both boundaries and internal regions within semantic parts (ref. Fig. \ref{fig:active_mask_comparison} bottom-left). Especially passing from 2.2\% to 5\% acquisition budget, HALO considers thick boundaries, so also parts of objects close to the boundaries, but also areas within objects, as it happens for wall, fence, pole, and sidewalk, cf.\ the right image part in the bottom-right of Fig. \ref{fig:active_mask_comparison}.
\end{itemize}

\end{document}